\documentclass[journal,twoside]{IEEEtran}

\ifCLASSINFOpdf
  \usepackage[pdftex]{graphicx}
  \graphicspath{{figures/}}
  \DeclareGraphicsExtensions{.pdf,.jpeg,.png}
\else
  \usepackage[dvips]{graphicx}
  \graphicspath{{figures/}}
  \DeclareGraphicsExtensions{.eps}
\fi

\usepackage{diagbox}
\usepackage{multirow}

\usepackage{cite}

\usepackage{color,xcolor}

\usepackage{amsmath}
\usepackage{amsfonts}

\usepackage{array}

\ifCLASSOPTIONcompsoc
  \usepackage[caption=false,font=normalsize,labelfont=sf,textfont=sf]{subfig}
\else
  \usepackage[caption=false,font=footnotesize]{subfig}
\fi
\usepackage{float}

\usepackage{threeparttable}

\usepackage{url}

\begin{document}
\title{A Time-Delay Feedback Neural Network for Discriminating Small, Fast-Moving Targets in Complex Dynamic Environments}

\author{Hongxin Wang, Huatian Wang, Jiannan Zhao, Cheng Hu, Jigen Peng and Shigang Yue, \IEEEmembership{Senior Member,~IEEE} \thanks{This work was supported in part by the National Natural Science Foundation of China under Grant 12031003 and Grant 11771347, in part by the European Union’s Horizon 2020 research and innovation programme under the Marie Sklodowska-Curie grant agreement No 691154 STEP2DYNA and No 778602 ULTRACEPT. \emph{(Corresponding authors: Jigen Peng; Shigang Yue.)}}
	\thanks{Ho. Wang, C. Hu and S. Yue are with the the Machine Life and Intelligence Research Center, Guangzhou University, Guangzhou 510006, China, and also with
		the Computational Intelligence Laboratory, School of Computer Science, University of Lincoln, Lincoln LN6 7TS, U.K. (email: howang@lincoln.ac.uk, syue@lincoln.ac.uk).}
	\thanks{Hu. Wang is with the Northwest Institute of Mechanical and Electrical Engineering, Xianyang 712099, China, and also with
		the Computational Intelligence Laboratory, School of Computer Science, University of Lincoln, Lincoln LN6 7TS, U.K..}
	\thanks{J. Zhao is with the the School of Electrical Engineering, Guangxi University, Nanning 530004, China, and also with
		the Computational Intelligence Laboratory, School of Computer Science, University of Lincoln, Lincoln LN6 7TS, U.K..}
	\thanks{J. Peng is with the School of Mathematics and Information Science, Guangzhou University, Guangzhou 510006, China (email: jgpeng@gzhu.edu.cn).}
}

\markboth{IEEE TRANSACTIONS ON Neural Networks and Learning Systems}
{Wang \MakeLowercase{\textit{et al.}}: A Time-Delay Feedback Neural Network for Discriminating Small, Fast-Moving Targets}
%



\maketitle

\begin{abstract}
Discriminating small moving objects within complex visual environments is a significant challenge for autonomous micro robots that are generally limited in computational power. By exploiting their highly evolved visual systems, flying insects can effectively detect mates and track prey during rapid pursuits, even though the small targets equate to only a few pixels in their visual field. The high degree of sensitivity to small target movement is supported by a class of specialized neurons called small target motion detectors (STMDs). Existing STMD-based computational models normally comprise four sequentially arranged neural layers interconnected via feedforward loops to extract information on small target motion from raw visual inputs. However, feedback, another important regulatory circuit for motion perception, has not been investigated in the STMD pathway and its functional roles for small target motion detection are not clear. In this paper, we propose an STMD-based neural network with feedback connection (Feedback STMD), where the network output is temporally delayed, then fed back to the lower layers to mediate neural responses. We compare the properties of the model with and without the time-delay feedback loop, and find it shows preference for high-velocity objects. Extensive experiments suggest that the Feedback STMD achieves superior detection performance for fast-moving small targets, while significantly suppressing background false positive movements which display lower velocities. The proposed feedback model provides an effective solution in robotic visual systems for detecting fast-moving small targets that are always salient and potentially threatening.  
\end{abstract}

\begin{IEEEkeywords}
Neural system modeling, visual neural systems, small target motion detection, time-delay feedback, complex background.
\end{IEEEkeywords}

%
\IEEEpeerreviewmaketitle

\section{Introduction}
\label{Introduction}
\begin{figure}[t!]
	\centering
	\includegraphics[width=0.45\textwidth]{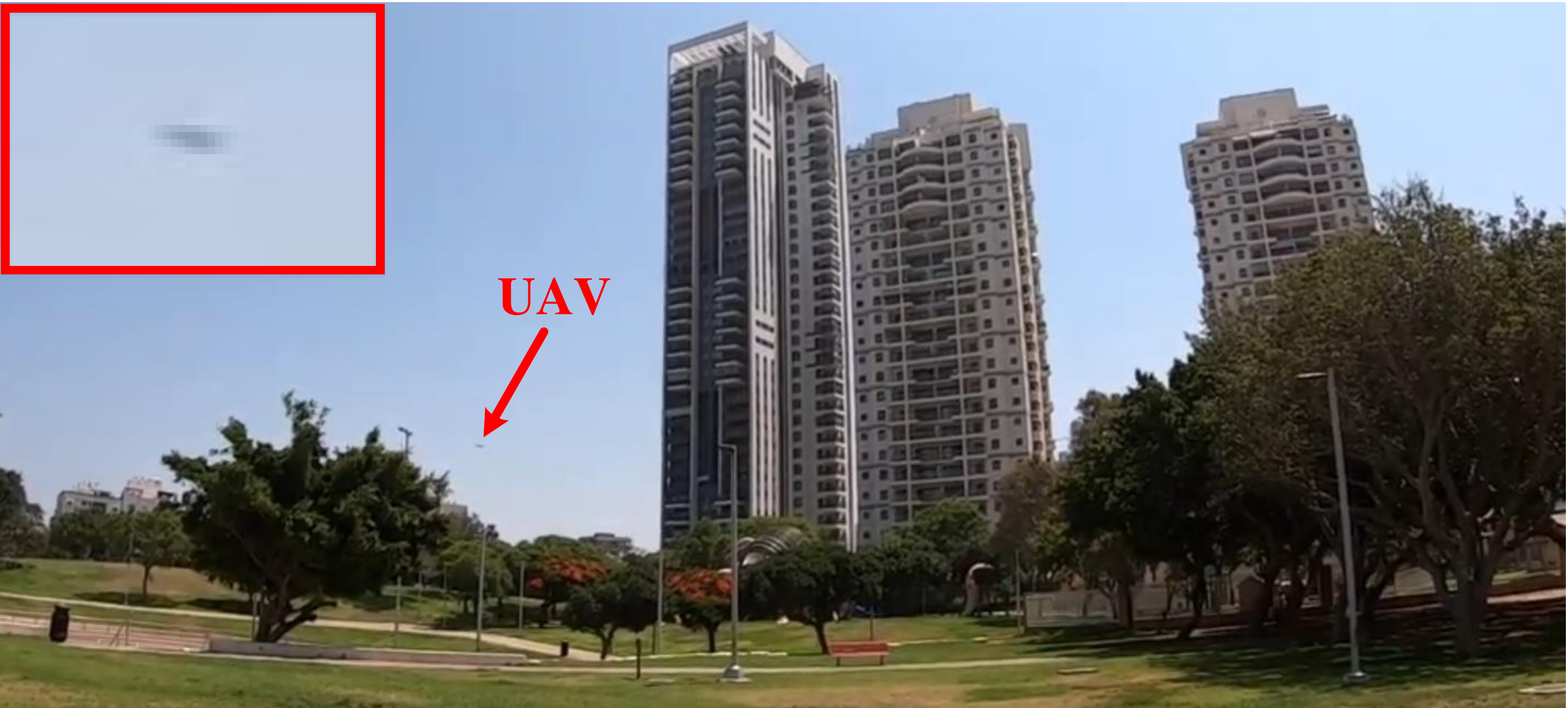}
	\caption{Example of a UAV in the distance where the UAV and its surrounding region are enlarged in the red box \cite{Youtube-UAV}.}
	\label{Schematic-UAV-in-theDistance}
\end{figure}

\IEEEPARstart{T}{he} projected images of real-world moving objects on a retina or a camera are likely to change significantly due to variation in distance, position, orientation, shape, occlusion, and lighting conditions \cite{yamins2014performance}. The ability to robustly detect visual motion of critical objects is important in the functioning of intelligent robots, particularly those performing monitoring and tracking tasks in complex dynamic environments \cite{wang2020bioinspired,sun2020decentralised,yue2013redundant,HuBin2016Rotation,yue2006collision}. To react promptly and maintain a dominant position in interaction/competition, artificial visual systems often need to detect moving objects as early as possible and at the greatest distance as possible, for example, in the early warning of incoming unmanned aerial vehicles (UAVs). The considerable initial distance of the objects from the visual sensors means that they usually appear as small dim speckles on images, only one or a few pixels in size, revealing almost no other visual features, as shown in Fig.  \ref{Schematic-UAV-in-theDistance}. Small target\footnote{Small targets refer to objects of interest that appear as small dim speckles on images due to long observation distance. Their size may vary from $1$ pixel to a few pixels, or equivalent to $1^{\circ}-3^{\circ}$ in visual field preferably.} motion detection aims not only to identify the locations and velocities of small targets, but also to discriminate them from other large, independently moving objects against heavily cluttered backgrounds.

Small target motion detection has a wide variety of real-world applications such as safe navigation in unknown complex environments, video surveillance over a wide area, and the warning of potential dangers in advanced autonomous driving. However, detecting small moving targets in complex dynamic environments is much more challenging than large object detection. Traditional motion detection methods \cite{javed2018moving,fortun2015optical,saleemi2013multiframe,Pedro2010Object,Anna2007Image,Redmon_2016_CVPR} perform well on large objects which permit visualization with a high degree of  resolution, and which present a clear appearance and structure, such as pedestrians, bikes, and vehicles. However, such methods are ineffective against targets as small as a few pixels. This is because visual features such as texture, color, shape, and orientation, are difficult to determine in such small objects and cannot be used for motion detection \cite{rozantsev2016detecting}. Moreover, small targets generally exhibit low resolution and unclear boundaries, which make them difficult to discriminate against cluttered backgrounds. Finally, free motion of the camera compounded with jittering may create further difficulties in motion discrimination. Effective solutions to detect small target motion against cluttered moving backgrounds on natural images are still rare.

Research in the field of visual neuroscience has contributed towards the design of artificial visual systems for small target detection \cite{fu2019towards,wang2019angular,liu2018event,hu2016bio,fu2018shaping}. As a result of millions of years of evolution, insects has developed accurate, efficient and robust capabilities for the detection of small moving targets. For example, dragonflies can track and intercept small flying prey or mates with limited neural resources and achieve an extremely high $97\%$ successful capture rate \cite{mischiati2015internal}. The high degree of insect sensitivity towards small target motion relies on a class of specialized neurons called small target motion detectors (STMDs) \cite{nordstrom2006insect,barnett2007retinotopic,nordstrom2012neural,wiederman2017predictive,kelecs2017object,nicholas2018integration}. STMD neurons are strongly excited by small targets occupying between $1^{\circ}-3^{\circ}$ of the visual field. In contrast, the neural responses to large objects (typically occupying $>10^{\circ}$) or background movements represented by wide-field grating stimuli, are much weaker or fall to spontaneous levels. In addition, the STMD neurons respond robustly to small targets against highly complex environments even in the presence of background motion. Understanding the basis for these superior neural properties is of  particular interest to researchers developing artificial visual systems for the efficient and robust detection of small target movement.

Examples of proposed quantitative STMD-based models include Elementary STMD (ESTMD) \cite{wiederman2008model}, Directionally selective STMD (DSTMD) \cite{wang2018directionally}, cascaded models \cite{wiederman2013biologically,bagheri2017performance,bagheri2017autonomous}, and STMD Plus \cite{wang2019Robust}. They each feature a feedforward processing hierarchy to transform raw visual inputs into strong responses to small moving targets. Despite the success of these feedforward models in small target motion detection, they are unable to completely filter out background false positives, particularly in complex dynamic environments. Accumulated evidence suggests that feedback is also critical to the visual perception processes in animals \cite{kietzmann2019recurrence, mohsenzadeh2018ultra,tang2019effective}. While feedforward loops can convey visual signals, feedback circuits modulate feature extraction in lower layers according to prior knowledge and the internal state \cite{bastos2015visual}. Specifically, they can exert positive and/or negative control over lower-layer neurons to enhance neural responses, and suppress distracting signals simultaneously \cite{clarke2017feedback,huang2018feedback}. 

Feedback mechanisms have been shown to be effective in improving model performance for a number of computer vision tasks, such as saliency detection \cite{li2018contrast}, pose estimation \cite{carreira2016human}, object recognition \cite{han2018deep}, and visual segmentation \cite{cao2018feedback}. Biological research has also identified various feedback loops in the visual systems of insects \cite{schnell2014cellular,karasek2018tailless,rosner2019neuronal,paulk2014selective}.  However, it has not been as deeply explored in STMD modeling of small target motion detection. In this paper, we propose an STMD-based model with time-delay feedback (Feedback STMD), and demonstrate its critical role in detecting small targets against cluttered backgrounds. We have conducted systematic analysis as well as extensive experiments. The results show that the Feedback STMD largely suppresses slow-moving background false positives, whilst retaining the ability to respond to small targets with higher velocities. The behaviour of the developed feedback model is consistent with that of the animal visual systems in which high-velocity objects always receive more attention \cite{gulyas1987suppressive,paulk2015closed,crowe2019goal}. Furthermore, it also enables autonomous robots to effectively discriminate potentially threatening {\bf fast-moving small targets} from complex backgrounds, a feature required, for example, in surveillance.

A preliminary version of our work has been presented in \cite{wang2018feedback}. In comparison, this paper presents several improvements: 1) we introduce the weighted summation of the neighboring STMD outputs as part of the feedback signal to suppress perturbation in the model output; 2) we provide a more comprehensive description and analysis of the functional mechanism of the Feedback STMD model; 3) we support the analysis with a parameter sensitivity study on the feedback loop; 4) we report the results of more extensive experiments that illustrate the role of feedback in small target motion detection.

The rest of this paper is organized as follows: Section \ref{Related-Work} provides an overview of the current related research on motion detection and feedback mechanisms. Section \ref{Formulation-of-the-System} presents the proposed Feedback STMD model. Section \ref{Results-and-Discussions} reports the experimental results as well as performance comparisons with  existing models on both synthetic and real-world data sets. Finally, Section \ref{Conclusion} concludes this paper and highlights potential directions for future work.

\section{Related Work}
\label{Related-Work}

In this section, we first review the STMD-based models, then describe feedback mechanism and its applications, finally discuss traditional motion detection approaches.

\subsection{STMD-based Models}
STMDs \cite{nordstrom2006insect,barnett2007retinotopic,nordstrom2012neural,wiederman2017predictive,kelecs2017object,nicholas2018integration} are a class of widely investigated motion-sensitive neurons that respond most strongly to small moving targets. Several attempts have been made to develop the STMD-based models for small target motion detection. Wiederman \emph{et al.} \cite{wiederman2008model} proposed a computational ESTMD model, to simulate size selectivity of STMD neurons. The presence of a small moving target is detected by the correlation of laterally inhibited luminance-change signals at each pixel. However, the ESTMD model does not show directional selectivity and cannot determine direction of motion. To address these issues, two directionally selective models have been  developed, including DSTMD \cite{wang2018directionally} and cascaded models \cite{wiederman2013biologically,bagheri2017performance,bagheri2017autonomous}. In these, directional selectivity is introduced by correlating signals from two different pixels while motion direction is estimated by a population vector algorithm. Wang \emph{et al.} \cite{wang2019Robust} proposed an STMD Plus model to explore the combination of motion information with directional contrast for filtering out small-target-like features. The above computational models all process visual signals in a feedforward manner. However, feedback loops which significantly outnumber feedforward connections in the visual systems of insects \cite{tuthill2016mechanosensation}, have not been systematically investigated in STMD neural modeling.

\subsection{Feedback Mechanism}

Feedback is a ubiquitous regulatory circuit in insects' visual systems, which mediates lower-layer neural responses by bringing back higher-level semantic information \cite{schnell2014cellular}. For example,  fruit flies are capable of fine tuning motor behavior using visual feedback during the pursuit of mates or when tracking prey \cite{karasek2018tailless}; praying mantids deliver binocular disparity feedback to the optic lobes to modulate stereo vision \cite{rosner2019neuronal}; visual selective attention mechanisms of honeybees involve feedback from the central brain \cite{paulk2014selective}. 

Over the past decade, feedback mechanisms have been successfully incorporated into artificial neural networks to accomplish a variety of visual tasks, including saliency detection \cite{li2018contrast}, pose estimation \cite{carreira2016human}, object recognition \cite{han2018deep}, and visual segmentation \cite{cao2018feedback}. In addition, they have also been used extensively in control theory to solve the stability problem of nonlinear systems \cite{li2019neural,ding2018neural,liu2018neural}. Although feedback mechanisms have achieved great success in improving performance of systems, little work has been done to model them in the STMD neural pathways and their role in small target motion detection remains unclear.

\subsection{Traditional Motion Detection Methods}

Traditionally, methods of motion detection have often been categorized into temporal differencing \cite{javed2018moving}, optical flow \cite{fortun2015optical}, and background subtraction \cite{saleemi2013multiframe}. Temporal differencing methods calculate the pixel-wise differences between consecutive frames to detect moving objects. Optical flow methods utilize the brightness constancy constraint to estimate distribution of apparent velocities of moving objects. Background subtraction methods first estimate a reference image in the absence of moving objects (the background image), then identify moving regions by assessing the differences between the current frame and the background image. These methods perform well against static backgrounds. However, their performance decline significantly in noisy and/or dynamic backgrounds. In particular, they suffer from a large number of false positives in the presence of background motion, as small moving targets can be hidden among pixel errors and/or background noise when applying background motion compensation \cite{ren2003motion}. In addition, these methods are not sensitive to size, which means they cannot eliminate large moving objects. 

Appearance-based approaches \cite{Pedro2010Object,Anna2007Image,Redmon_2016_CVPR} can also be adopted for motion detection. They operate by extracting low-to-high level visual features to classify moving objects using machine learning algorithms, such as convolutional neural networks, random forests, and deformable part models. These methods have excellent detection performance for objects that are sufficiently large and with discriminative visual features in individual images. Nevertheless, they usually fail to detect small targets moving against cluttered backgrounds. This is because the visual features are difficult to discern due to the poor-quality appearance of  extremely small targets \cite{rozantsev2016detecting}. 

In previous research efforts, majority of small target detection methods focus on infrared images \cite{deng2021infrared,bai2018derivative,gao2013infrared}, which are heavily dependent on significant temperature differences between the background and small targets, such as rockets, jets, and missiles.  In addition, the detection environment of the infrared-based methods is mainly sky or ocean that is quite clear and homogeneous on infrared images. Although these methods perform well on infrared images, they are powerless for small target motion detection on natural images against cluttered backgrounds with lots of bushes, trees, sunlight and shadows.

\section{Methods and Formulations}
\label{Formulation-of-the-System}

\begin{figure}[t!]
	\centering
	\includegraphics[width=0.48\textwidth]{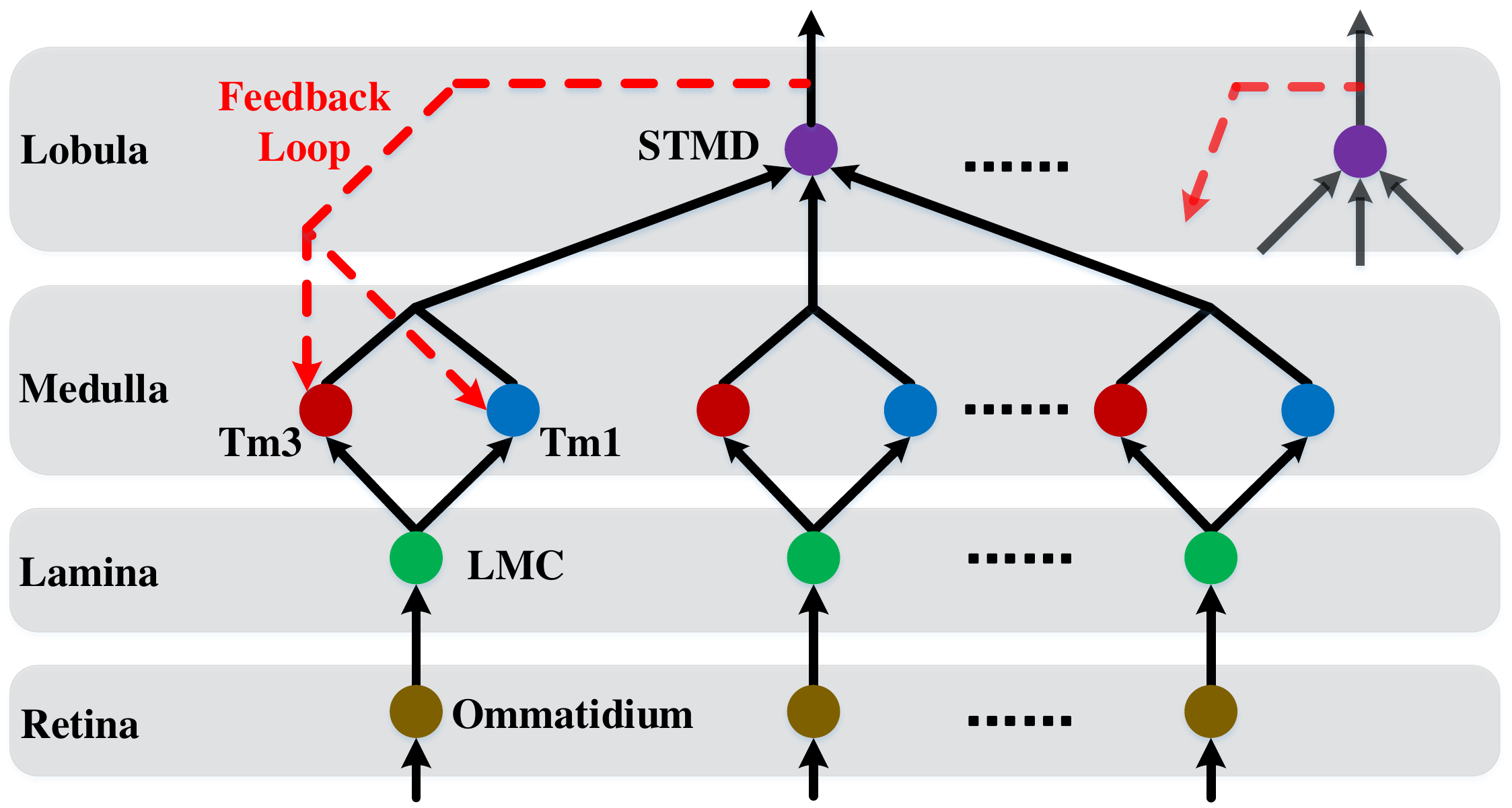}
	\caption{Wiring sketch of the proposed Feedback STMD model. The model consists of four neural layers: the retina, lamina, medulla and lobula (from bottom to top). Each layer contains a number of specific neurons denoted by colored circular nodes. The STMDs interact with the medulla neurons (Tm1 and Tm3) through feedforward (black lines) and feedback (red lines) loops. Note that only one feedback loop is presented here for clarity.}
	\label{Schematic-of-Visual-System-Feedback-Loop}
\end{figure}

\begin{figure*}[t!]
	\centering
	\includegraphics[width=0.90\textwidth]{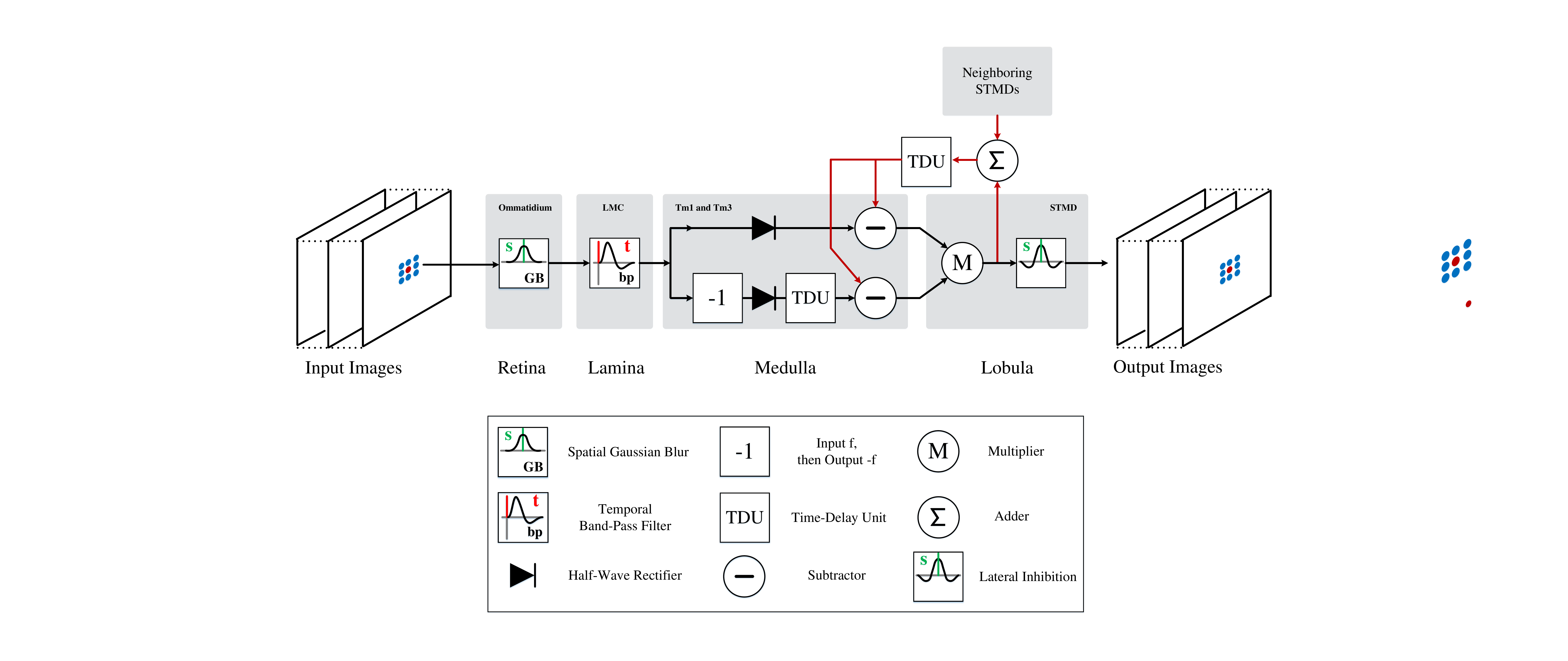}
	\caption{Schematic diagram of the proposed Feedback STMD model. At each time step, the model takes an entire image frame as input and extracts motion information concerning small targets via four feedforward neural layers. Initially, the extracted motion information is temporally delayed, then fed back to the medulla layer to inhibit neural responses to slow-moving objects. We show only one ommatidium, LMC, Tm1, Tm3, and STMD here for clarity. However,  they are arranged in matrix form in corresponding neural layers.}
	\label{Schematic-of-Feedback-STMD-Model}
\end{figure*}

The proposed Feedback STMD model is composed of four sequentially arranged neural layers -- the retina, lamina, medulla, and lobula, as illustrated in Fig. \ref{Schematic-of-Visual-System-Feedback-Loop}. Each layer contains a number of specialized visual neurons coordinated to detect small target motion against dynamic complex environments. Specifically, visual information is received by ommatidia \cite{meglivc2019horsefly}, then fed into large monopolar cells (LMCs) \cite{behnia2014processing} to calculate changes in luminance over time. The output from the LMCs is further processed by medulla neurons (Tm1 and Tm3) \cite{takemura2013visual} in parallel, and finally integrated in the STMDs to discriminate small moving targets. The STMDs propagate their outputs to the medulla neurons via feedback loops to mediate neural responses to various visual stimuli. Fig. \ref{Schematic-of-Feedback-STMD-Model} shows the schematic of the proposed Feedback STMD model; its formulation will be described in the following subsections.

\begin{figure*}[t!]
	\centering
	\includegraphics[width=1\textwidth]{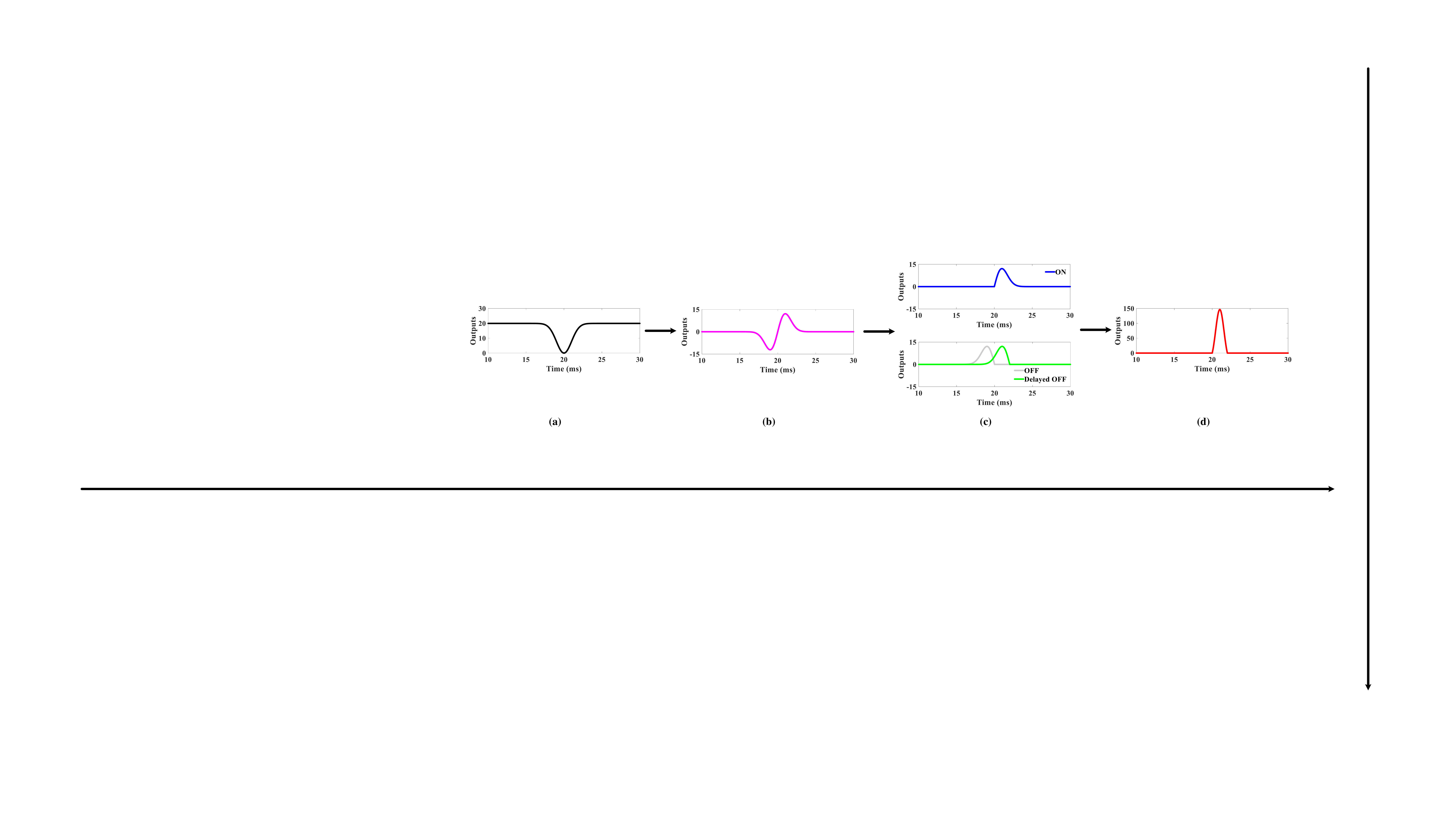}
	\caption{Neural outputs to a dark small target with respect to time $t$ at a given pixel $(x_0,y_0)$. (a) Ommatidium output $P(x_0,y_0,t)$. (b) LMC output $L(x_0,y_0,t)$. (c) Tm3 output $S^{Tm3}(x_0,y_0,t)$ and Tm1 output $S^{Tm1}(x_0,y_0,t)$. (d) STMD output $D(x_0,y_0,t)$.}
	\label{Schematic-of-Signal-Processing-ESTMD}
\end{figure*}

\subsection{Retina Layer}
The retina layer is built from thousands of individual ommatidia that serve as luminance receptors to capture visual information from the external environment. In the proposed model, the ommatidia are arranged in matrix form to receive an entire image frame as the input [see Fig. \ref{Schematic-of-Feedback-STMD-Model}]. Each neuron is designed as a Gaussian filter in spatial domain to smooth the luminance signal of each pixel. Formally, we denote the input image sequence as $I(x,y,t) \in \mathbb{R}$, where $x,y$ and $t$ stand for spatial and temporal coordinates, respectively. The output of an ommatidium $P(x,y,t)$ is defined by the convolution of $I(x,y,t)$ with a Gaussian function $G_{\sigma_1}(x,y)$, namely, 
\begin{align}
P(x,y,t) &=  \iint I(u,v,t)G_{\sigma_1}(x-u,y-v)dudv  \\
G_{\sigma_1}(x,y)& = \frac{1}{2\pi\sigma_1^2}\exp(-\frac{x^2+y^2}{2\sigma_1^2})
\label{Photoreceptors-Gaussian-Blur}
\end{align}
where $\sigma_1$ is the standard deviation of the Gaussian function.

To visualize the signal processing of the proposed model clearly, we illustrate the neural outputs that result from a dark small moving target in Fig. \ref{Schematic-of-Signal-Processing-ESTMD}. When the small target passes through pixel $(x_0,y_0)$, the output of the ommatidium first declines between $15$ ms and $20$ ms, then gradually goes up to the original level during the next period ($20$ to $25$ ms), as shown in Fig. \ref{Schematic-of-Signal-Processing-ESTMD}(a). It should be pointed out that the decrease and increase of the ommatidium output are induced by the arrival and departure of the small target, respectively.

\subsection{Lamina Layer}
As can be seen from Fig. \ref{Schematic-of-Feedback-STMD-Model}, the output of the ommatidia form the input to the LMCs in the lamina layer, each of which is modeled as a band-pass filter to compute luminance change at each pixel with respect to time. Considering the desirable features of Gamma kernel in temporal processing, such as trivial stability, easy adaptation and the uncoupling of the region of support of the impulse response and the order \cite{principe1993gamma,de1991theory}, we define the impulse response of the temporal band-pass filter $H(t)$ by the difference of two Gamma kernels, that is,
\begin{align}
H(t) &= \Gamma_{n_1,\tau_1}(t) - \Gamma_{n_2,\tau_2}(t) \label{BPF-Para}\\
\Gamma_{n,\tau}(t) &= (nt)^n \frac{\exp(-nt/\tau)}{(n-1)!\cdot \tau^{n+1}}
\end{align}
where $\Gamma_{n,\tau}(t)$ denotes Gamma kernel with order $n$ and time constant $\tau$. By convolving $H(t)$ with the ommatidium output $P(x,y,t)$, we have the output of the LMC $L(x,y,t)$
\begin{equation}
L(x,y,t) = \int P(x,y,s)H(t-s) ds.
\label{LMCs-BPF}
\end{equation}
The LMC output $L(x,y,t)$ reveals the change of luminance corresponding to pixel $(x,y)$ at time $t$. A positive output reflects an increase in luminance whereas a negative one corresponds to a decrease in luminance, as shown in Fig. \ref{Schematic-of-Signal-Processing-ESTMD}(b). It is important to note that the LMC is unable to discriminate between objects that differ in size. Hence, the LMC output that indicates changes in luminance could result from the motion of an object of any-size. To filter out large moving objects, the output of the LMC is fed-forward to higher neural layers for further processing.

\subsection{Medulla Layer}
Two medulla neurons, including Tm1 and Tm3, directly connect with the LMC and process its output in parallel, as illustrated in Fig. \ref{Schematic-of-Feedback-STMD-Model}. Specifically, the Tm3 neuron is modelled as a half-wave rectifier to pass luminance increase component while block decrease component. Note that the luminance increase and decrease components are also referred to as ON and OFF signals, respectively. By denoting the output of Tm3 by $S^{\text{Tm3}}(x,y,t)$, then we have
\begin{equation}
S^{\text{Tm3}}(x,y,t) = [L(x,y,t)]^{+} 
\end{equation}
where $[x]^+$ represents $\max (x,0)$. In contrast, the Tm1 neuron passes on the decreasing luminance component and further temporally delays it by convolution with a Gamma kernel, that is,
\begin{equation}
S^{\text{Tm1}}(x,y,t) = \int [-L(x,y,s)]^{+} \cdot  \Gamma_{n_3,\tau_3}(t-s) ds \label{Tm1-Output}
\end{equation}
where $S^{\text{Tm1}}(x,y,t)$ denotes the output of Tm1. The time-delay length and time-delay order are determined by the time constant $\tau_3$ and order $n_3$ of Gamma kernel $\Gamma_{n_3,\tau_3}(t)$, respectively. 

As can be seen from Fig. \ref{Schematic-of-Signal-Processing-ESTMD}(c), the outputs of the Tm3 (ON) and Tm1 (OFF) are aligned correctly in the temporal field after applying the time delay. The length of the time-delay is set as the time taken for the small target to pass through the pixel, i.e., the ratio of the target width to its velocity. The aligned ON and OFF signals are further multiplied together to produce a large response to the small moving target [see Fig. \ref{Schematic-of-Signal-Processing-ESTMD}(d)]. In addition to the feedforward processing, the ON- and OFF-type cells may optimize neural coding for object motion using feedback from higher neural layers, as revealed in recent studies \cite{clarke2017feedback,huang2018feedback}. Inspired by this, the proposed model exerts feedback control over the medulla neurons to regulate their responses to different moving objects, as displayed in Fig. \ref{Schematic-of-Feedback-STMD-Model}. To simplify the explanation, the whole feedback loop is described in the next subsection.

\subsection{Lobula Layer} 
\label{Lobula-Layer}

\begin{figure}[t!]
	\centering
	\includegraphics[width=0.45\textwidth]{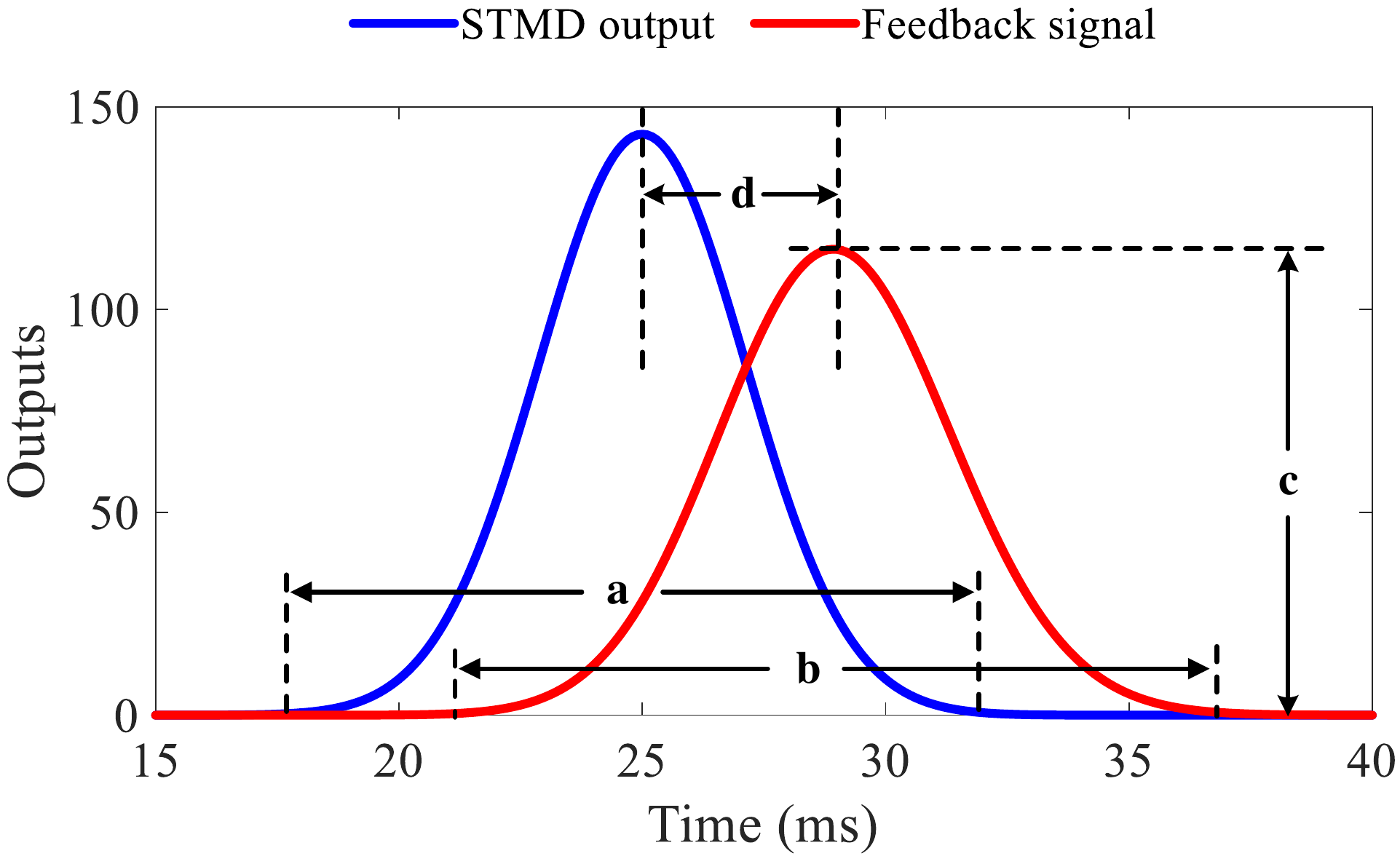}
	\caption{The STMD output and feedback signal with respect to time $t$ at a given pixel $(x_0,y_0)$. Letter a represents the response duration of the STMD output, while letters b, c, and d denote the response duration, the strength, and time-delay length of the feedback signal, respectively.}
	\label{Schematic-Model-Output-Feedback-Signal}
\end{figure}

It can be observed from Fig. \ref{Schematic-of-Feedback-STMD-Model} that each STMD integrates the two medulla neural outputs from the same pixel during small target motion detection. More precisely, a feedback signal is initially subtracted from the medulla neural outputs, which are then recombined by multiplication to generate a significant response, that is,
\begin{equation}
\begin{split}
D(x,y,t) =&  \Big\{ S^{\text{Tm3}}(x,y,t) -  F(x,y,t) \Big\} \\  & \times 
\Big\{ S^{\text{Tm1}}(x,y,t) -  F(x,y,t)\Big\}
\label{ESTMD-Signal-Correlation}
\end{split}
\end{equation}
where $D(x,y,t)$ denotes the output of the STMD neuron, and $F(x,y,t)$ represents the feedback signal. Note that the feedback signal is always associated with time delays due to time taken for conduction along the axon or dendrite and/or time taken for transmission across the synapse. To model the time delay in a feedback loop, the outputs of the central STMD and its neighbors are temporally delayed by convolving with a Gamma kernel, then propagated to the medulla layer as feedback signals, that is,  
\begin{equation}
F(x,y,t) = \alpha \cdot \int \Big\{D(x,y,s) + E(x,y,s) \Big\}\cdot \Gamma_{n_4,\tau_4}(t-s) ds 
\label{Feedback-Signal-Delay} 
\end{equation}
where $\alpha$ is a feedback constant, $E(x,y,t)$ stands for the weighted summation of the neighboring STMD neural outputs which is used to suppress signal perturbation [see Supplementary Materials],  and $n_4$ and $\tau_4$ are the order and time constant of the Gamma kernel $\Gamma_{n_4,\tau_4}(t)$, respectively. Here, we define the weight function as
\begin{equation}
W_e(x,y)= \frac{1}{2\pi\eta^2}\exp(-\frac{x^2+y^2}{2\eta^2})
\label{Feedback-Loop-Weight-Function-of-Surrounding-STMDs}
\end{equation}
where $\eta$ is a constant which is determined by the preferred size range of the STMD to ensure that the weight function $W_e(x,y)$ can cover the STMD neural outputs in local spatial region elicited by the small target motion. Then, $E(x,y,t)$ can be given by
\begin{equation}
\begin{split}
E(x,y,t) = \iint \Big\{S^{\text{Tm3}}(&u,v,t) \times S^{\text{Tm1}}(u,v,t)\Big\} \\
                &\times W_e(x-u,y-v) dudv.
\end{split}
\end{equation}

\begin{figure}[t!]
	\centering
	\subfloat[]{\includegraphics[width=0.45\textwidth]{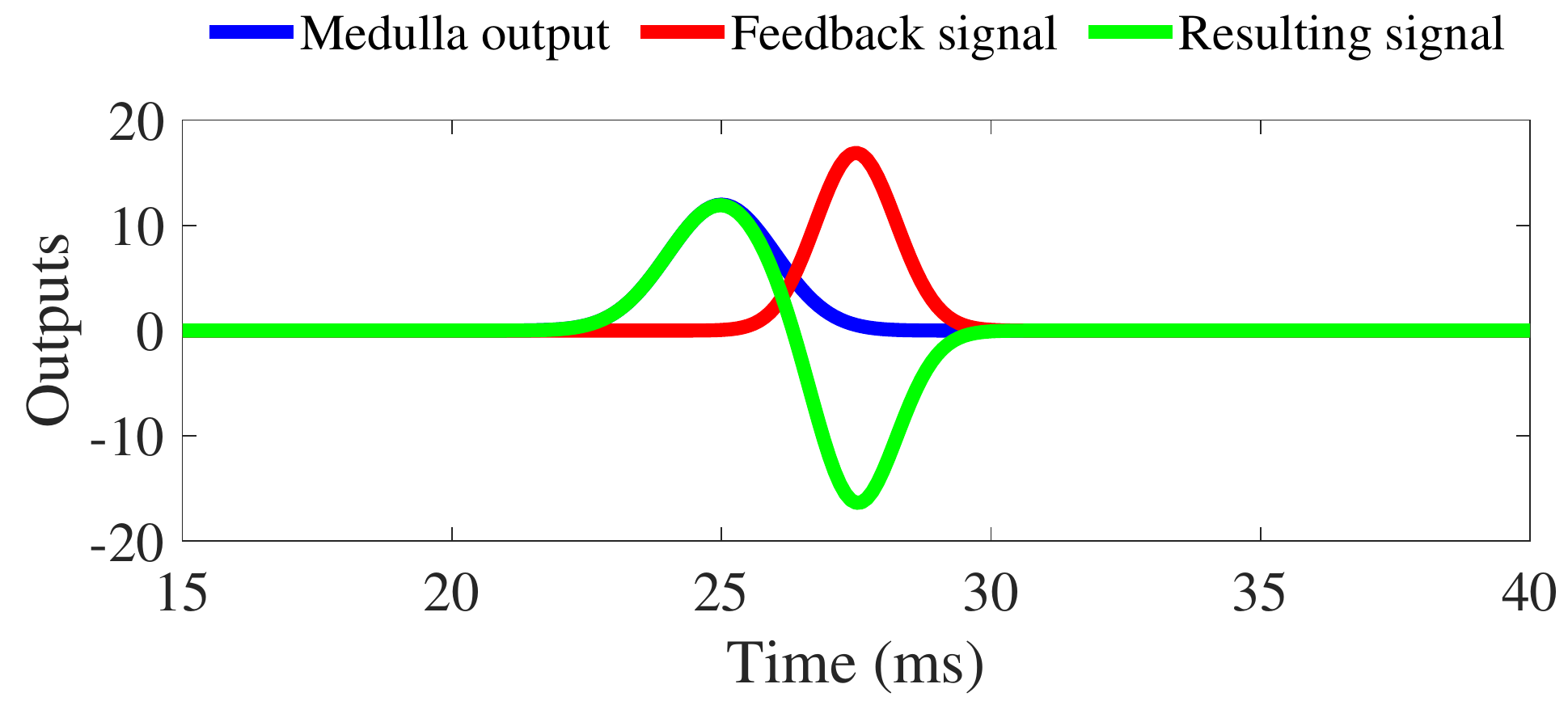}}
	\hfil
	\subfloat[]{\includegraphics[width=0.45\textwidth]{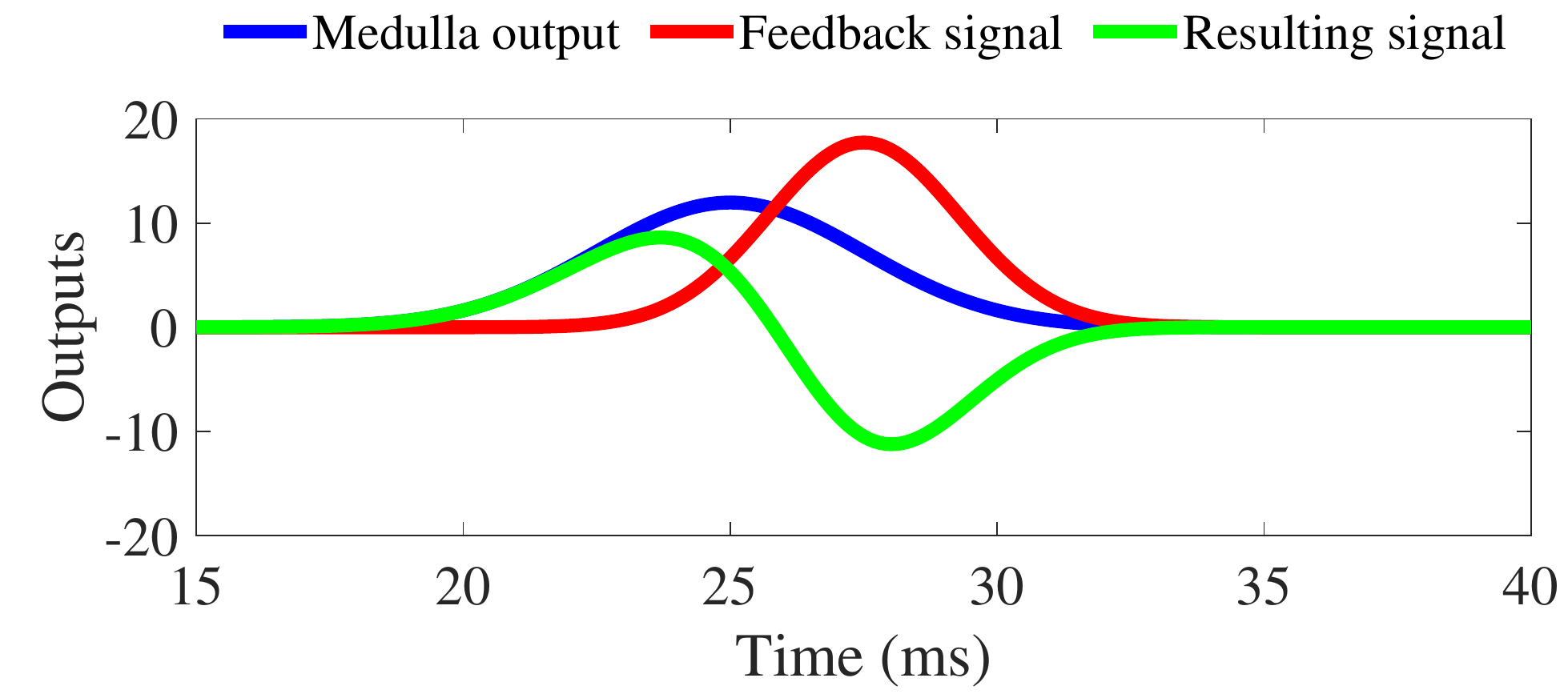}}
	\caption{Feedback to medulla neurons with (a) short response duration, and (b) long response duration, where the feedback constant,  order, and time-delay length of the feedback loop are fixed. Note that a longer response duration is indicative of a lower object velocity. The medulla output with short response duration maintains at its maximum after feedback, whereas the long-response-duration output is strongly suppressed.} 
	\label{Schematic-Medulla-Signal-Feedback-Signal-Result}
\end{figure}

\begin{figure}[t!]
	\centering
	\subfloat[]{\includegraphics[width=0.45\textwidth]{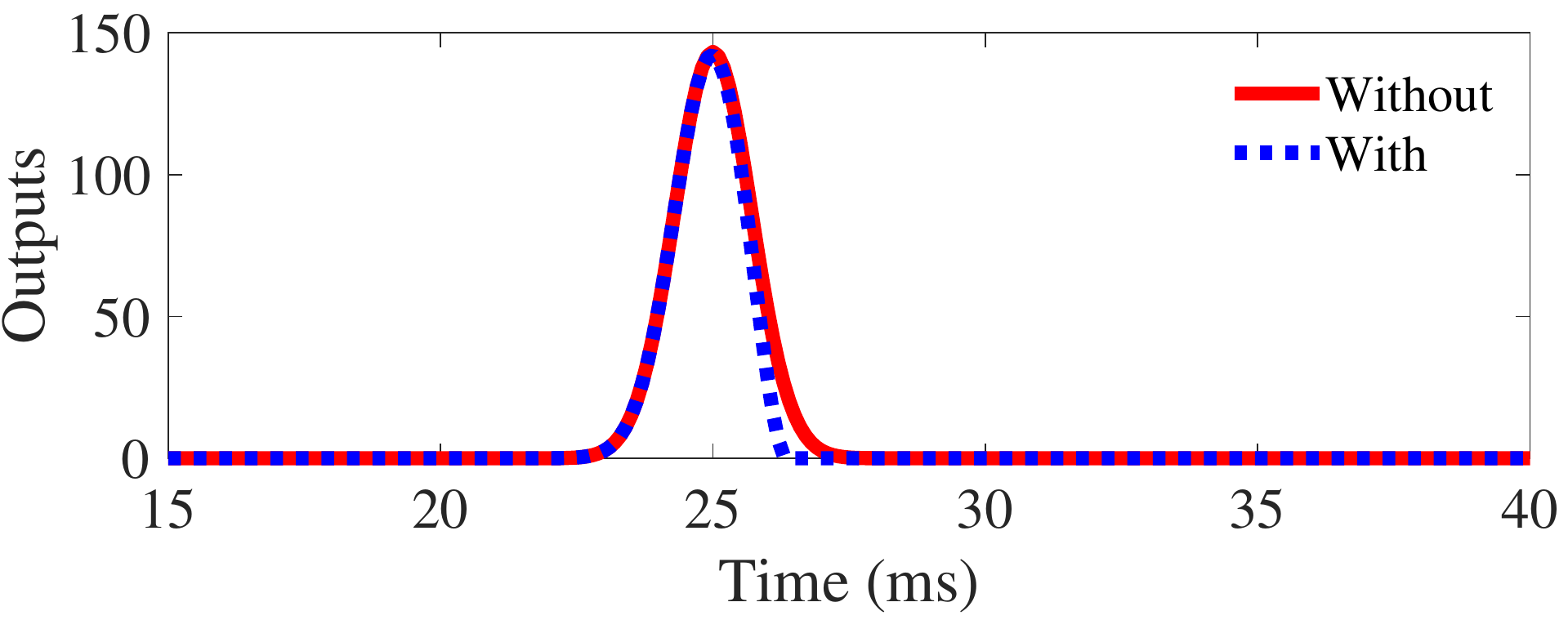}}
	\hfil
	\subfloat[]{\includegraphics[width=0.45\textwidth]{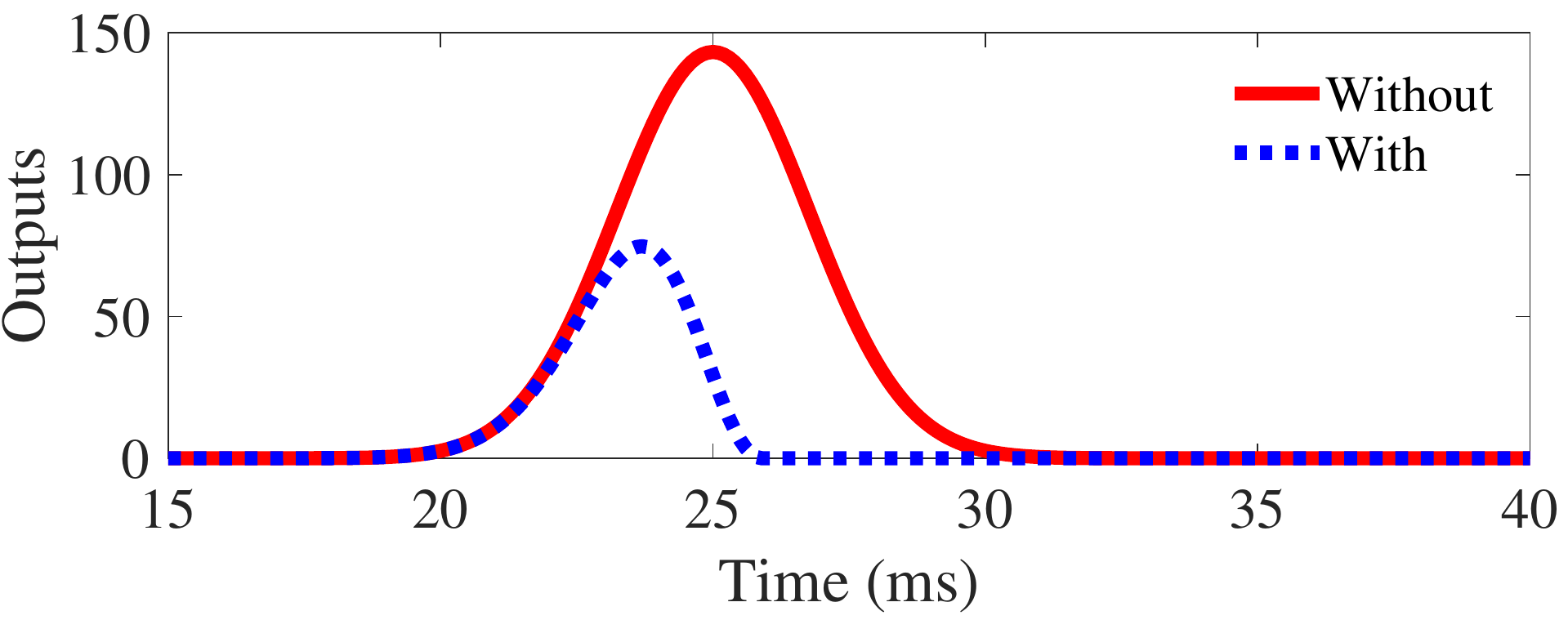}}
	\caption{Outputs of the STMD with and without feedback. (a) Short response duration. (b) Long response duration. The STMD output with short response duration maintains its maximum after feedback, whereas the long-response-duration output is strongly suppressed.}
	\label{Schematic-STMD-Feedback-STMD-Results}
\end{figure}

\textbf{\textit{Analysis of the Time-Delay Feedback:}} To demonstrate the role of the time-delay feedback, we first analyse the STMD output and its feedback signal. As illustrated in Fig. \ref{Schematic-Model-Output-Feedback-Signal}, the STMD output is significantly greater than zero during a specific time period referred to as the response duration\footnote{Response duration represents the time elapsing between onset of the response and its termination, i.e., recovering to the original level.}.  The response duration of the STMD is determined by the time taken for the object to completely cover the pixel, i.e., the  reciprocal of the object's velocity $1/v$. A longer response duration means that more time is taken for the object to cover the pixel, and is indicative of a lower object velocity. Here we formulate this relation as: $\text{a}=f(1/v)$  where $f(\cdot)$ is an increasing function. The feedback signal is the delayed version of the STMD output, where its response duration, strength, and time-delay length are controlled by the parameters $n_4$, $\alpha$, and $\tau_4$, respectively [see (\ref{Feedback-Signal-Delay}) and Fig. \ref{Schematic-Model-Output-Feedback-Signal}]. Moreover, the response duration of the feedback signal is larger than that of the STMD (i.e., $\text{b}>\text{a}$), according to the convolution property.

We further compare neural outputs with different response durations after applying the negative feedback, as shown in Figs. \ref{Schematic-Medulla-Signal-Feedback-Signal-Result} and \ref{Schematic-STMD-Feedback-STMD-Results}. Obviously, if the time-delay length is greater than half of the response duration of the feedback signal, i.e., $\text{d}>\text{b}/2$, the outputs from the medulla neurons and the STMD output will maintain their maxima which will remain unchanged after subtracting the feedback signal [see Fig. \ref{Schematic-Medulla-Signal-Feedback-Signal-Result}(a) and \ref{Schematic-STMD-Feedback-STMD-Results}(a)]. Thus, we have
\begin{align}
\text{d} & >\frac{\text{b}}{2}>\frac{\text{a}}{2} = \frac{f(1/v)}{2} \\
v & > \frac{1}{f^{-1}(\text{b})} >\frac{1}{f^{-1}(2\text{d})}
\label{velocity-threshold}
\end{align}
where $f^{-1}(\cdot)$ is the inverse function of $f(\cdot)$. These two equations suggest that when the time-delay length is fixed, the feedback loop has a minor effect on the neural responses to objects with velocities larger than $1/f^{-1}(2\text{d})$; on the contrary, if $v < 1/{f^{-1}(2\text{d})}$, the medulla neural outputs and the STMD output will all be significantly suppressed by the feedback signal [see Fig. \ref{Schematic-Medulla-Signal-Feedback-Signal-Result}(b) and \ref{Schematic-STMD-Feedback-STMD-Results}(b)]. Note that $1/{f^{-1}(2\text{d})}$ is determined by the time-delay feedback signal and can be tuned by the three feedback parameters $\alpha$, $n_4$, and $\tau_4$, which will be further discussed in Section \ref{Parameter-Sensitivity-Analysis}.  The above analysis indicates that the time-delay feedback loop shows a preference for fast-moving objects\footnote{Fast-moving objects refer to those whose velocities are higher than $1/f^{-1}(2\text{d})$.}, which is consistent with the biological finding that visual systems of animals generally pay more attention to fast-moving objects than those moving at lower velocities \cite{gulyas1987suppressive,paulk2015closed,crowe2019goal}.

At this stage, the STMD model can eliminate slow-moving objects by the proposed time-delay feedback, while continuing to respond to those with higher velocities regardless of their size. To suppress responses to large objects, we further convolve the STMD output $D(x,y,t)$ with a lateral inhibition kernel $W_s(x,y)$, namely, 
\begin{equation}
Q(x,y,t) = \iint D(u,v,t) W_s(x-u,y-v) du dv
\label{DS-STMD-Lateral-Inhibition}
\end{equation}
where $Q(x,y,t)$ denotes the laterally inhibited output, $W_s(x,y)$ is defined as
\begin{align}
W_s(x,y) &= A \cdot [g(x,y)]^{+} + B \cdot [g(x,y)]^{-}  \label{Inhibition-Kernel-W2-1}\\
g(x,y)  &= G_{\sigma_2}(x,y) - e \cdot G_{\sigma_3}(x,y) - \rho
\label{Lateral-Inhibition-Kernel-W2-2}
\end{align}
where $[x]^+$ and $[x]^-$ represent $\max (x,0)$ and $\min (x,0)$, respectively, $A$, $B$, $e$ and $\rho$ are constant. To identify the locations of small targets, the model output $Q(x,y,t)$ is compared with a detection threshold $\lambda$. If $Q(x,y,t) > \lambda$, the location $(x,y)$ is considered as a positive detection. In other words, we believe that a small target is detected at the location $(x,y)$.

\section{Experimental Results and Discussions}
\label{Results-and-Discussions}

\subsection{Experimental Setup}
\subsubsection{Data Sets}
We have evaluated our proposed Feedback STMD model using two publicly available data sets, including Vision Egg \cite{straw2008vision} and RIST \cite{RIST-Data-Set}. The Vision Egg data set includes a number of synthetic image sequences, each of which displays a computer generated small target (i.e., a black block) moving against real background images. The synthetic videos cover a wide variety of  background and target types with different parameters, such as luminance, velocity, and size. Their sampling frequency is set to $1000$ Hz, while resolution is $500$ pixels (horizontal) by $250$ pixels (vertical). The RIST data set consists of $16$ videos captured in real-world environments using an action camera (GoPro Hero $6$) at $240$ fps. Each video holds an object with size ranging between $3\times 3$ and $15 \times 15$ pixels, and contains various types of challenging scenarios, such as a highly cluttered background, a low contrast object, sudden camera motion, and bad weather conditions.

\subsubsection{Evaluation Criteria}
We employed detection rate and false alarm rate to quantitatively evaluate the model performance. The two metrics can be calculated as
\begin{align}
D_R & = \frac{\text{number of true positives}}{\text{number of actual targets}} \\
F_A & = \frac{\text{number of false positives}}{\text{number of images}}
\end{align}
where $D_R$ and $F_A$ denote detection and false alarm rates, respectively. A detection is counted as true positive if its distance to the ground truth is smaller than a threshold ($5$ pixels).

\begin{table}[t]
	\renewcommand{\arraystretch}{1.3}
	\caption{Parameters of the Proposed Feedback STMD Model}
	\label{Table-Parameter-Feedback-STMD}
	\centering
	\begin{tabular}{cc}
		\hline
		Eq. & Parameters \\	
		\hline
		(\ref{Photoreceptors-Gaussian-Blur}) & $\sigma_1 = 1$ \\
		
		(\ref{BPF-Para}) & $n_1 = 4, \tau_1= 8, n_2 = 16,\tau_2 = 32$\\
		
		(\ref{Tm1-Output}) &  $n_3 = 9, \tau_3= 45$ \\
		
		(\ref{Feedback-Signal-Delay})    &  $\alpha = 1, n_4 = 10, \tau_4= 25$ \\ 
		
		(\ref{Feedback-Loop-Weight-Function-of-Surrounding-STMDs})  & $\eta=1.5$ \\ 
		
		(\ref{Inhibition-Kernel-W2-1}) & $A = 1, B = 3$ \\
		
		(\ref{Lateral-Inhibition-Kernel-W2-2}) & $\sigma_2 = 1.5, \sigma_3 = 3, e = 1, \rho = 0$ \\                                  
		\hline
	\end{tabular}
\end{table}	

\subsubsection{Implementation}

We implemented the proposed Feedback STMD model in the Matlab environment on a laptop with a $2.20$GHz Intel Core i7 CPU and $16$GB memory. The parameters of the four feed-forward layers are determined by the analysis in \cite{wang2018directionally}. The three feedback parameters including the feedback constant $\alpha$, the order $n_4$  and time constant $\tau_4$ of the Gamma kernel, control the feedback signal as well as the velocity threshold $1/f^{-1}(2d)$, as shown in Section \ref{Lobula-Layer}. Tuning these three parameters will change $1/f^{-1}(2d)$, and also affect the model's preferred velocity range. To further evaluate the effect of the three parameters, we have conducted  a sensitivity study which is reported in Section  \ref{Parameter-Sensitivity-Analysis}. 

The parameter settings for the experimental results are given in Table \ref{Table-Parameter-Feedback-STMD}. Note that the given parameters will make the model show preference for fast-moving small objects. In unknown environments, multiple STMDs with different preferred velocity and size ranges could be coordinated to detect small objects.

\begin{figure}[!t]
	\centering
	\includegraphics[width=0.40\textwidth]{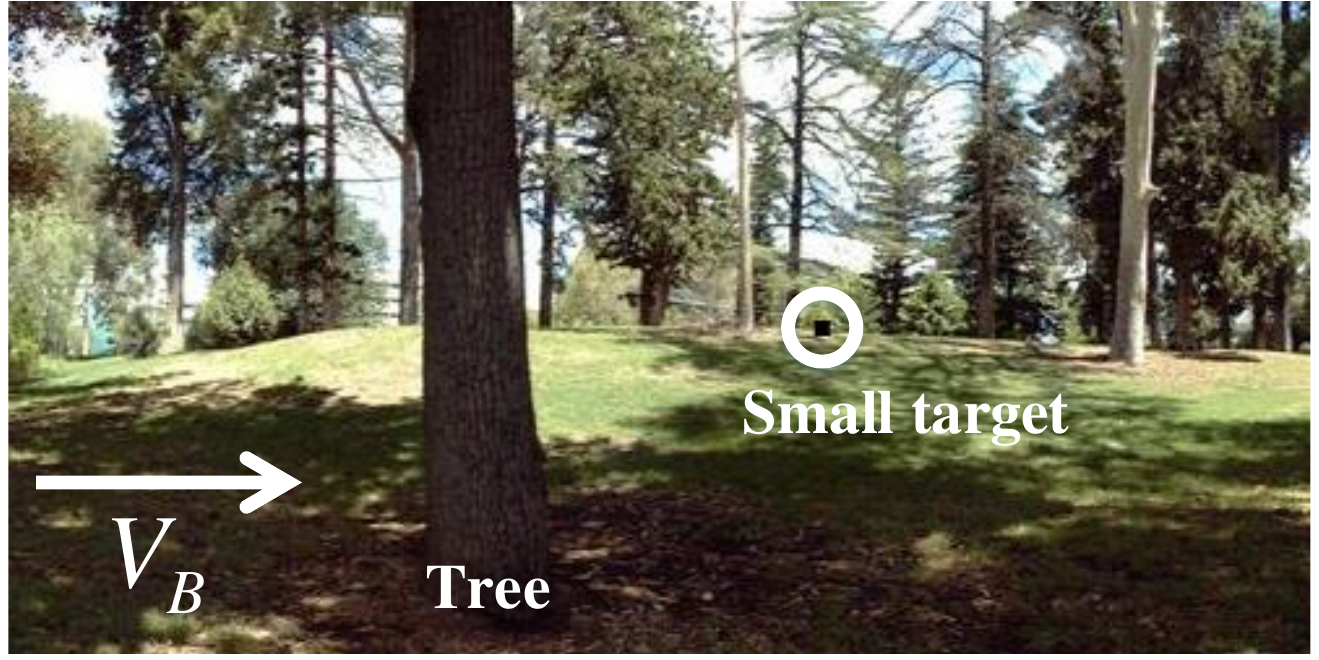}
	\caption{Input image at time $t_0 = 750$ ms where a small target (the black block) is moving against the cluttered natural  background. The velocity of the small target is set to $250$ pixels/s, while that of the background is lower ($150$ pixels/s). Arrow $V_B$ denotes the motion  direction of the background. The tree is considered as a large object moving with the background at the same velocity.}
	\label{Input-Image-Frame-Middle-Line-Highlighted}
\end{figure}

\begin{figure}[!t]
	\centering
	\includegraphics[width=0.48\textwidth]{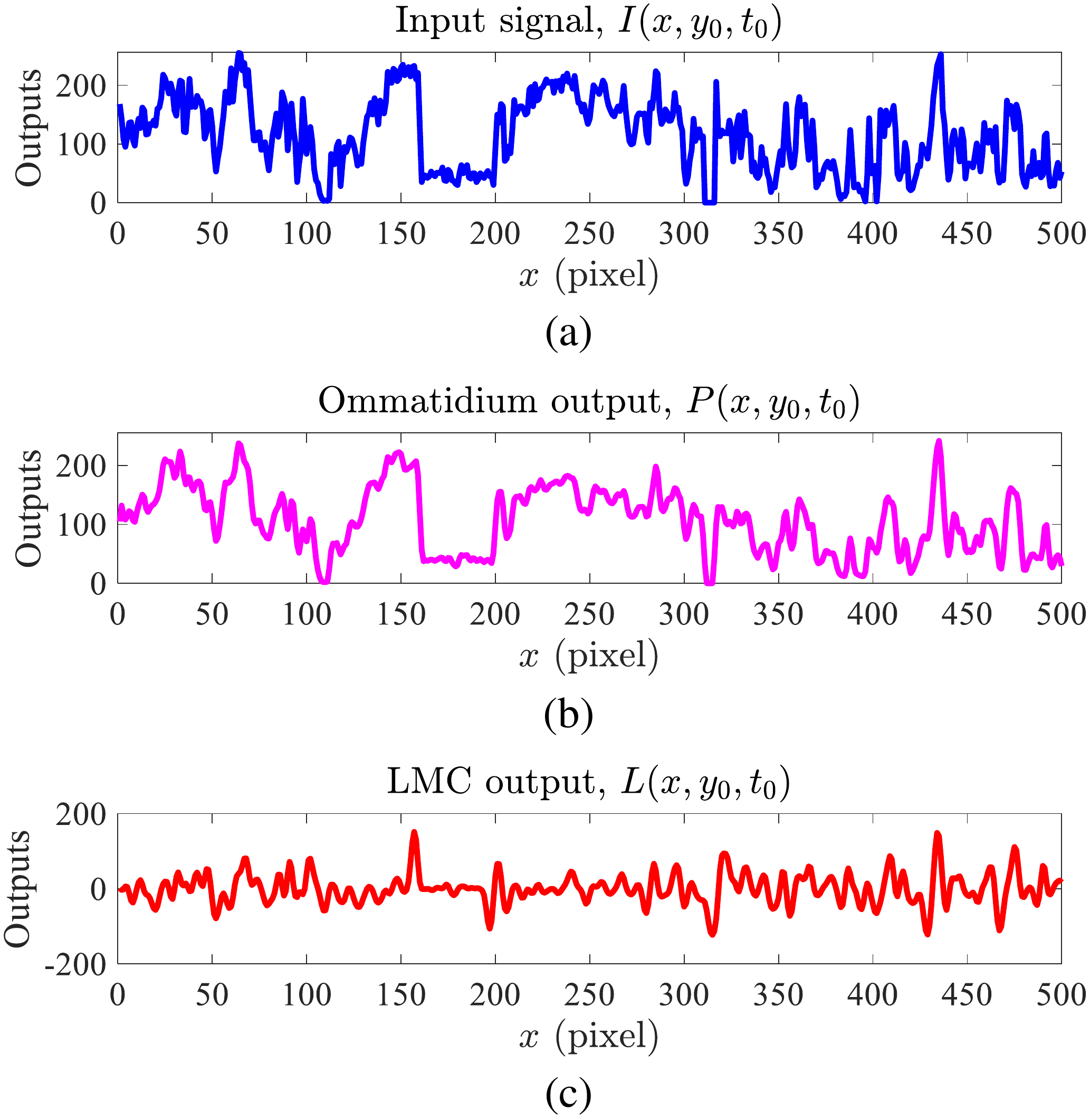}
	\caption{Neural outputs with respect to $x$ for the given $y_0=125$ pixels and time $t_0=750$ ms. (a) Input luminance signal $I(x,y_0,t_0)$. (b) Ommatidium output $P(x,y_0,t_0)$. (c) LMC output $L(x,y_0,t_0)$.}
	\label{Layer-Output-Input-Ommatidium-LMC}
\end{figure}

\begin{figure*}[!t]
	\centering
	\includegraphics[width=0.95\textwidth]{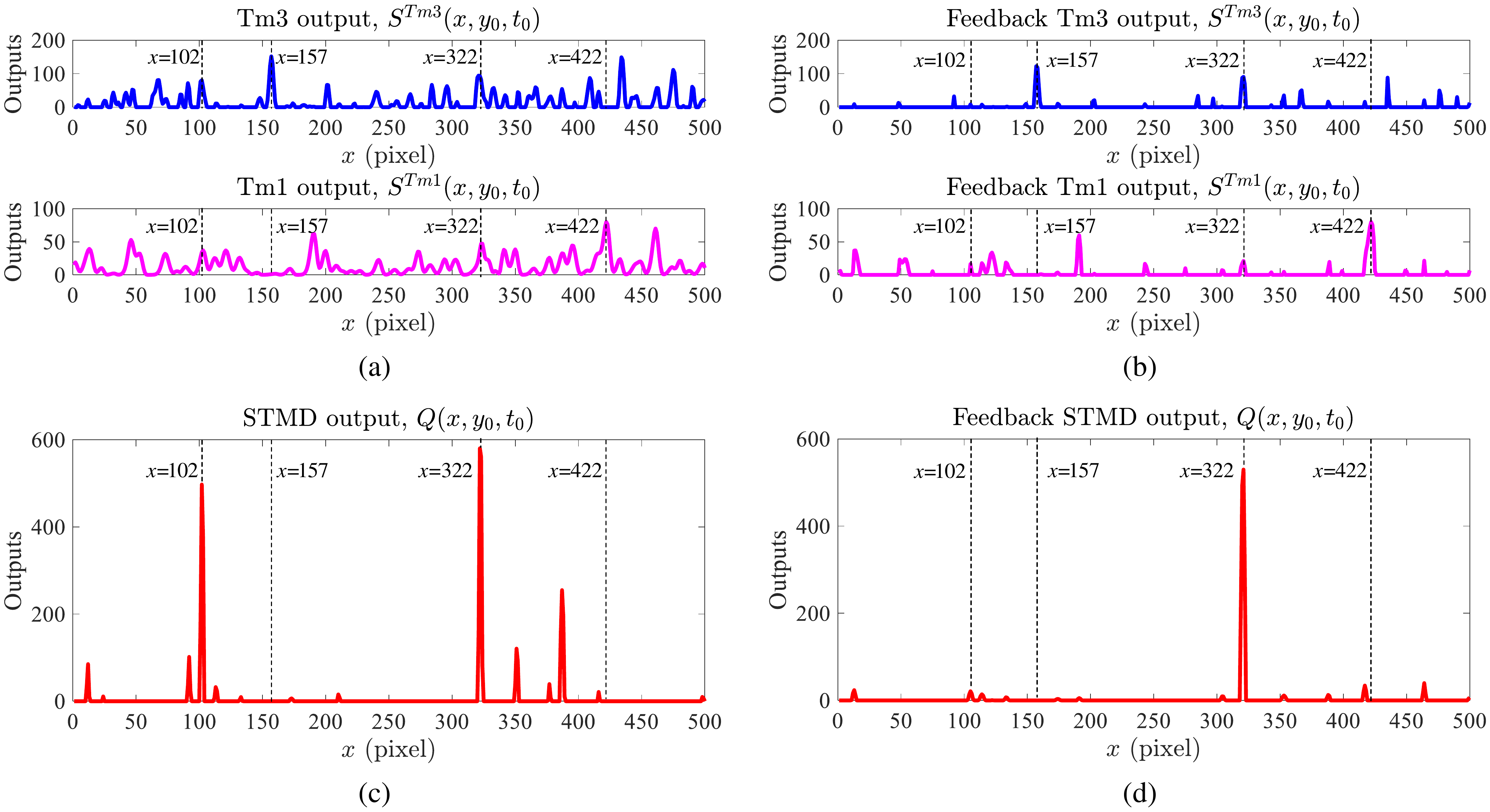}
	\caption{Comparison of neural outputs with and without feedback. Two medulla neural outputs (a) without and (b) with feedback, and the STMD output (c) without and (d) with feedback. The STMD model with feedback effectively suppresses responses to background false positives.}
	\label{Layer-Output-Medulla-STMD-Feedback-STMD}
\end{figure*}

\subsection{Effectiveness of the Time-Delay Feedback}

To validate the effectiveness of the time-delay feedback in suppressing slow-moving objects, we compare the neural outputs with and without feedback. Fig. \ref{Input-Image-Frame-Middle-Line-Highlighted} shows the input image at time $t_0 = 750$ ms where a small target is moving against a cluttered background. For better visualization and comparison of neural processing, we first set $y_0 = 125$ pixels then present the input luminance signal $I(x,y_0,t_0)$ with respect to $x$ as well as its resulting neural outputs in Fig. \ref{Layer-Output-Input-Ommatidium-LMC} and \ref{Layer-Output-Medulla-STMD-Feedback-STMD}. Since the time-delay feedback is only applied to the medulla layer, it does not affect the ommatidium output $P(x,y_0,t_0)$ and LMC  output $L(x,y_0,t_0)$, shown in Fig. \ref{Layer-Output-Input-Ommatidium-LMC}(b) and (c), respectively. As can be seen, 
the ommatidium output is a smooth blur resembling of the input luminance signal, whereas the LMC output reflects the luminance change over time at each pixel. Specifically, a positive LMC output at a pixel $x$ means luminance increases while a negative output represents a decrease in luminance.

We further compare the outputs of the medulla neurons and STMD neuron with and without feedback. Fig. \ref{Layer-Output-Medulla-STMD-Feedback-STMD}(a) displays the two medulla neural outputs without feedback, where the Tm3 output $S^{\text{Tm3}}(x,y_0,t_0)$ is the positive half of the LMC output while the Tm1 output $S^{\text{Tm1}}(x,y_0,t_0)$ is a time-delay version of the negative half. These two medulla neural outputs are multiplied together and then laterally inhibited to define the STMD neural output $Q(x,y_0,t_0)$ in Fig. \ref{Layer-Output-Medulla-STMD-Feedback-STMD}(c). It can be observed that the STMD exhibits a strong response at $x = 322$, i.e., the location of the small target, whereas the responses at other pixels such as $x = 157, 422$ are close to $0$. This is because the two maxima of the medulla neural outputs at $x=322$ are suitably aligned by the time delay [see Fig. \ref{Layer-Output-Medulla-STMD-Feedback-STMD}(a)], which finally results in a significant STMD output after multiplication and lateral inhibition. However, the outputs of the medulla neurons at the pixels $x = 157, 422$ are unable to achieve precise alignment, consequently leading to extremely low responses of the STMD neurons. Note that $x = 157$ corresponds to the location of the tree which is regarded as a large object in the input image. 

Nevertheless, we realize that the STMD model cannot completely filter out false positives against the cluttered background by signal alignment and lateral inhibition. As shown in Fig. \ref{Layer-Output-Medulla-STMD-Feedback-STMD}(c), the STMD still responds strongly at some background pixels, such as $x = 102$. The proposed time-delay feedback loop is able to suppress these background false positives utilizing differences in velocity between the small target and background. Since the background is moving at a lower velocity ($150$ pixels/s) compared to the small target ($250$ pixels/s), the medulla neural responses to the background features are strongly inhibited by subtracting the time-delay feedback signal, whereas the responses to the small target can be well maintained, as illustrated in Fig. \ref{Layer-Output-Medulla-STMD-Feedback-STMD}(b). Their resulting STMD output is shown in Fig. \ref{Layer-Output-Medulla-STMD-Feedback-STMD}(d). As can be seen, the STMD with feedback gives the maximal response to the small target ($x=322$), however, its responses to background features at other pixels are significantly suppressed.

In the above experiment, the time-delay feedback loop has demonstrated its ability to improve performance of the STMD model for small target motion detection by inhibiting slow-moving background features. Relative motion has been regarded as an important cue for animals to discriminate objects from cluttered environments \cite{kimmerle2000detection,cao2003neural,tadin2019spatial}. From this perspective, the proposed feedback loop provides a possible explanation for how relative motion information facilitates object discrimination against complex moving backgrounds.

\subsection{Tuning Properties of the Feedback STMD}

\begin{figure}[t]
	\centering
	\includegraphics[width=0.20\textwidth]{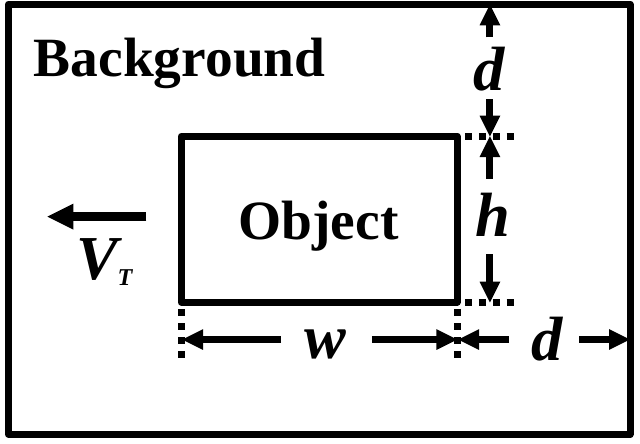}
	\caption{Illustration of an object and its neighboring background rectangle, where $V_T$, $w$, $h$, and $d$ denote motion direction of the object, object width, object height, and a constant, respectively.}
	\label{The-External-Rectangle-and-Neighboring-Background-Rectangle-of-a-Small-Target}
\end{figure}


\begin{figure}[t]
	\centering
	\subfloat[]{\includegraphics[width=0.23\textwidth]{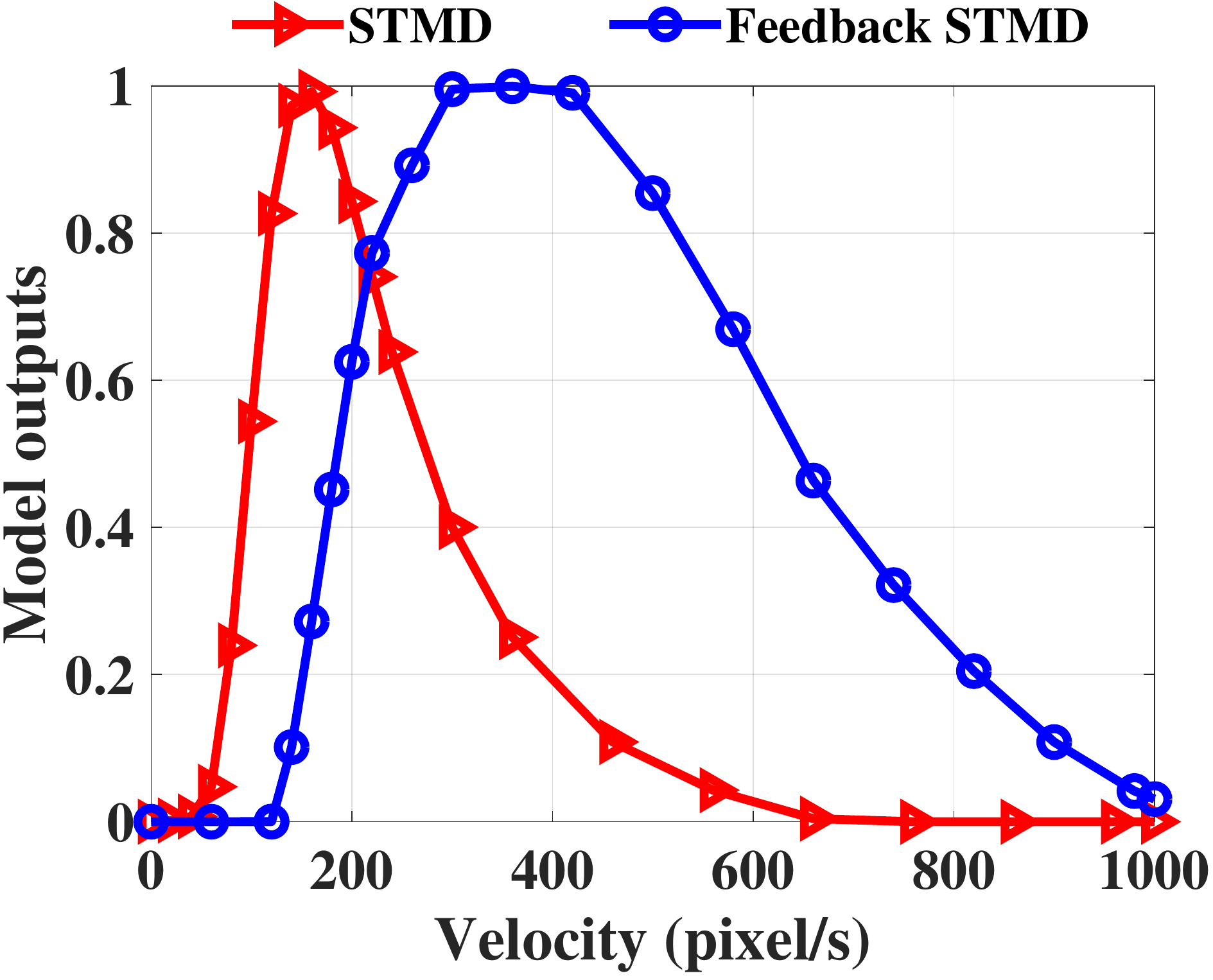}}
	\hfil
	\subfloat[]{\includegraphics[width=0.23\textwidth]{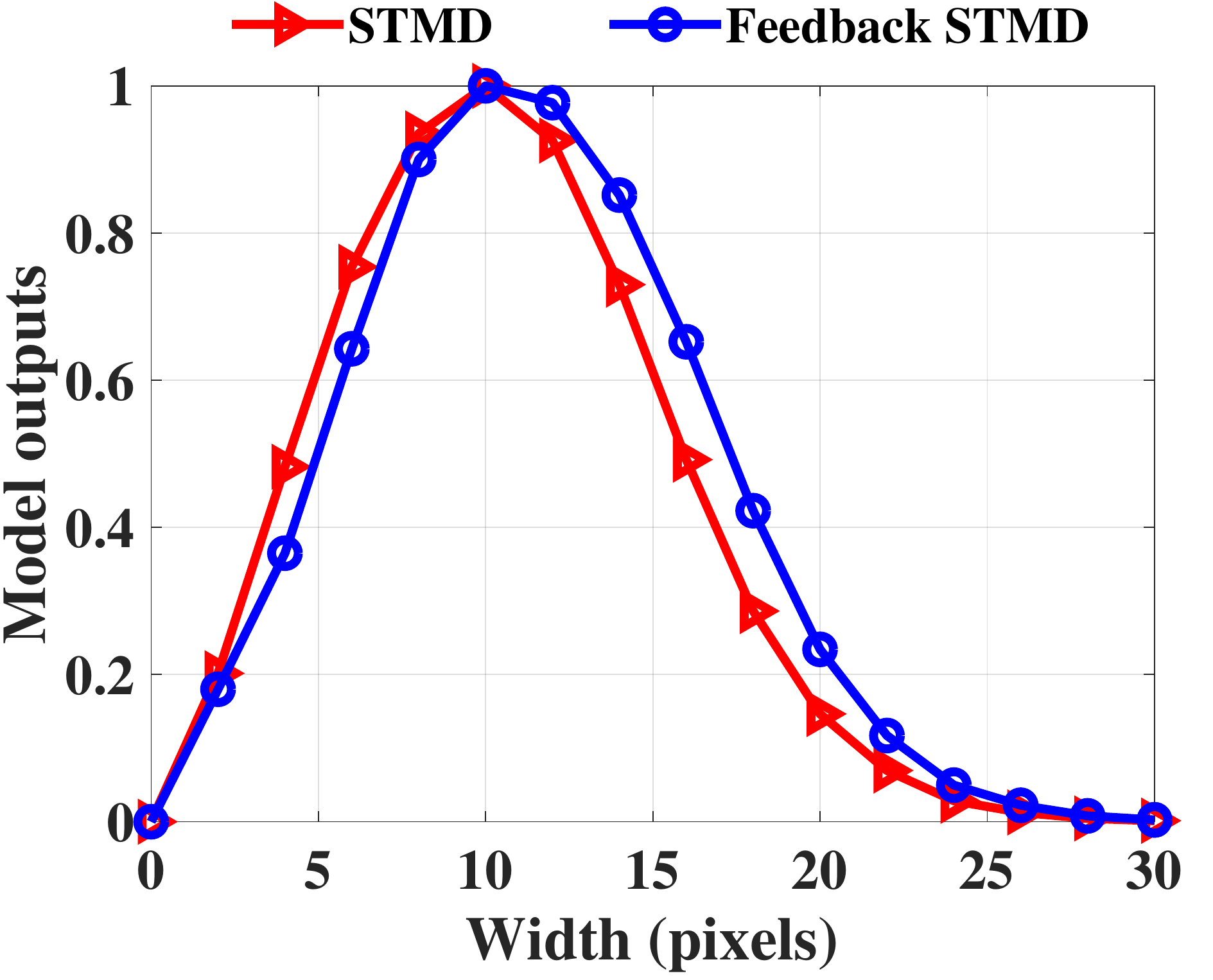}}
	\hfil
	\subfloat[]{\includegraphics[width=0.23\textwidth]{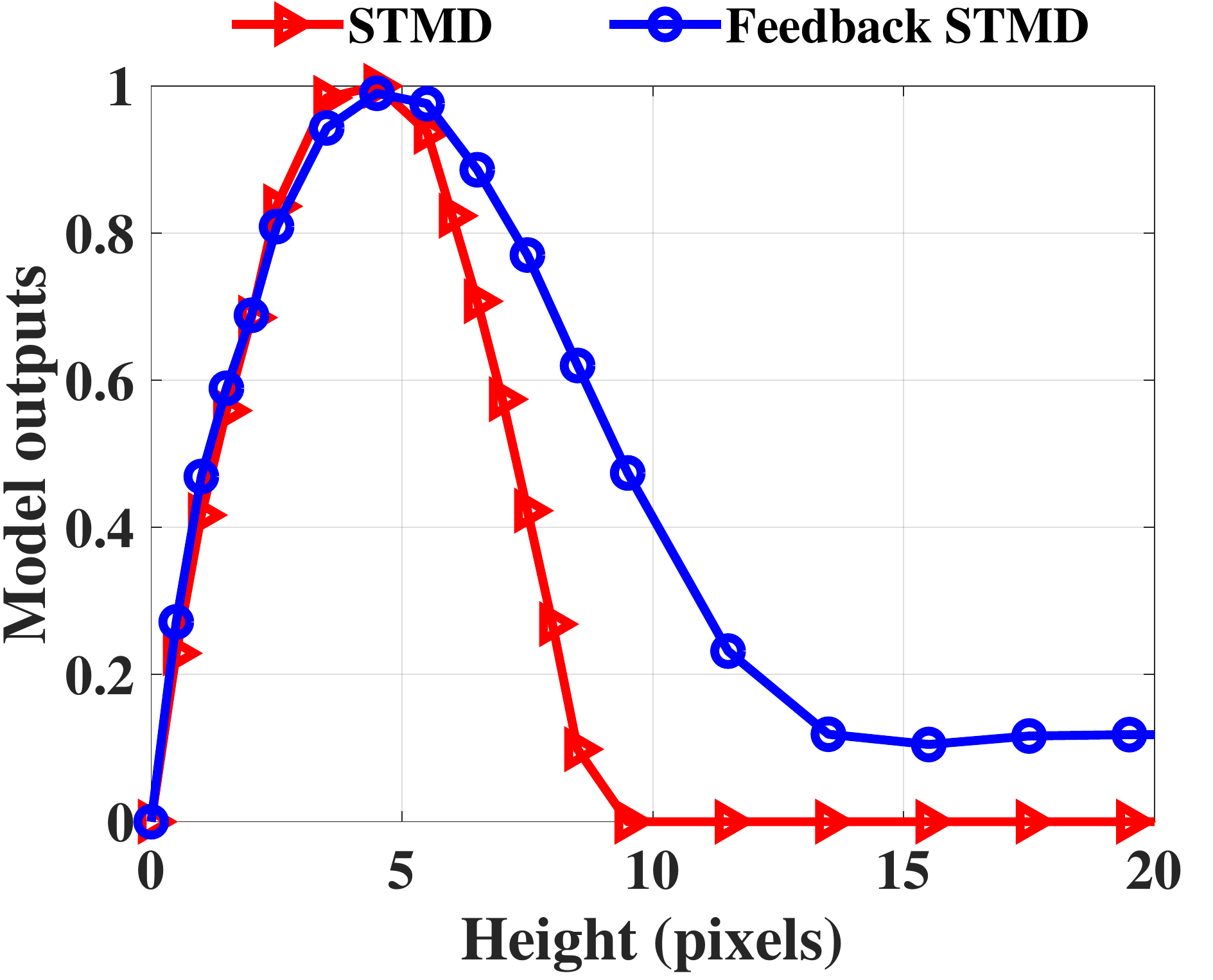}}
	\hfil
	\subfloat[]{\includegraphics[width=0.23\textwidth]{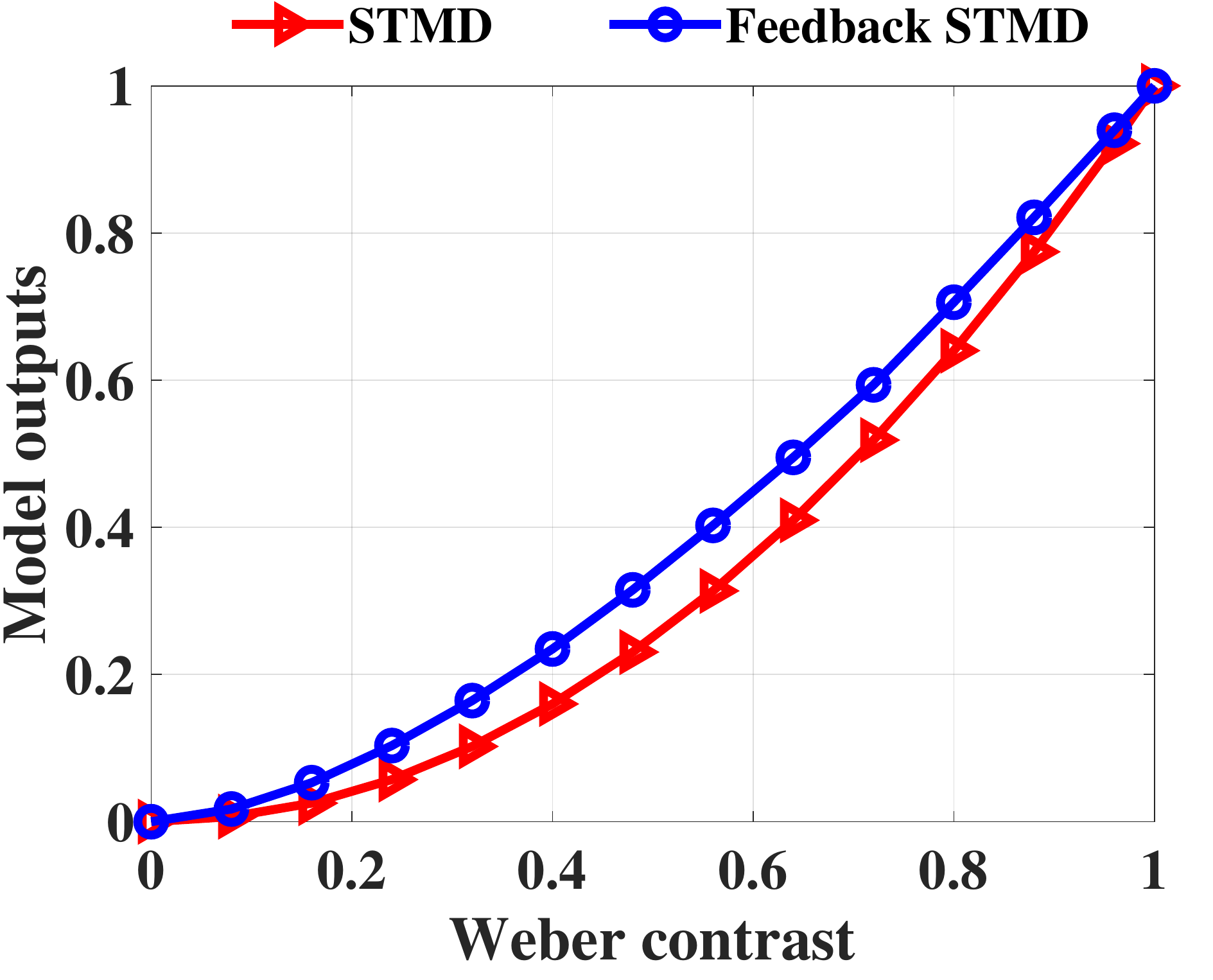}}
	\caption{Outputs of the STMD and Feedback STMD models to objects with different (a) velocities, (b) widths, (c) heights, and (d) Weber contrast. The two models all exhibit the four tuning properties, but have different optimal velocities and preferred velocity ranges.}
	\label{Tuning-Properties-Feedback-STMD}
\end{figure}

\begin{figure*}[t]
	\centering
	\includegraphics[width=1\textwidth]{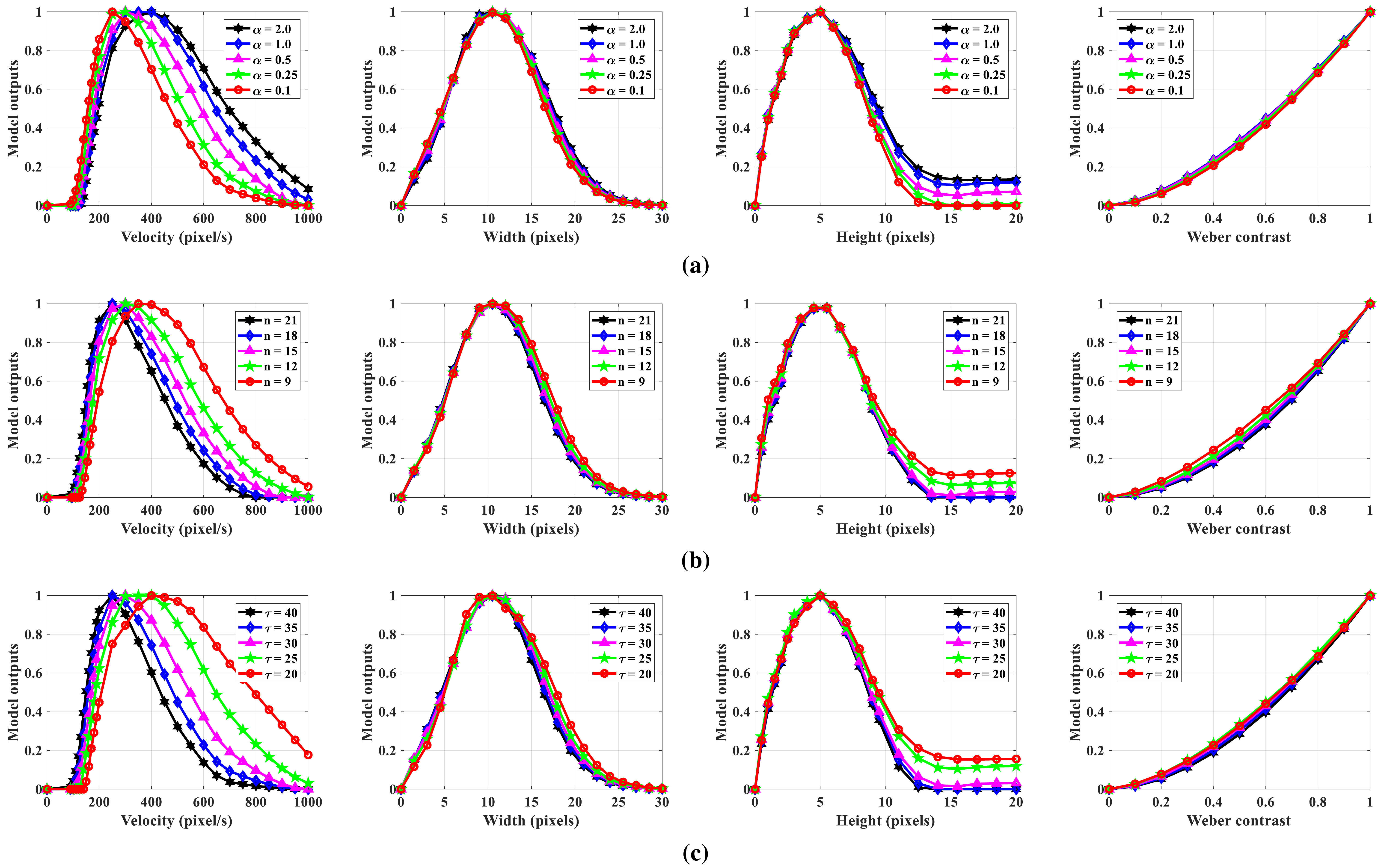}
	\caption{Tuning properties of the Feedback STMD for different values of (a) feedback constant $\alpha$, (b) time-delay order $n_4$, and (c) time-delay length $\tau_4$. Modifying the three parameters can significantly influence the optimal velocity and preferred velocity range, but have minor effects on width selectivity, height selectivity and Weber contrast sensitivity.}
	\label{Tuning-Properties-Different-Feedback-Alpha-Order-Tau}
\end{figure*}

As revealed in biological research \cite{nordstrom2006insect,barnett2007retinotopic,nordstrom2012neural,wiederman2017predictive,kelecs2017object,nicholas2018integration}, the STMD neurons display four distinct tuning properties including velocity selectivity, width selectivity, height selectivity, and Weber contrast sensitivity. To validate the tuning properties of the proposed model, we report the model outputs with respect to different object velocities, widths, heights, and Weber contrast in this subsection. For an object with size of $w \times h$ pixels, the size of its neighboring background rectangle is set as $(w+2d)\times(h+2d)$ where $d$ is a constant ($10$ pixels), as shown in Fig. \ref{The-External-Rectangle-and-Neighboring-Background-Rectangle-of-a-Small-Target}. Weber contrast measures the luminance difference between the object and its neighboring region, which can be calculated by
\begin{equation}
\text{Weber contrast} = \frac{|\mu_t - \mu_b|}{255}
\label{Weber-Contrast-Definition}
\end{equation}
where $\mu_t$ denotes average pixel intensity of the object, and $\mu_b$ represents average pixel intensity of the neighboring background region. We initialize the four parameters of the object, i.e., velocity, width, height, and Weber contrast, to $250$ pixels/s, $5$ pixels, $5$ pixels, and $1$, respectively, then record the model outputs by changing one of the object parameters while fixing the other three at their initial values.

Fig. \ref{Tuning-Properties-Feedback-STMD}(a)--(d) shows the outputs of the Feedback STMD model with respect to object velocity, width, height, and Weber contrast, respectively, where the outputs of the STMD (without feedback) is also provided for comparison. As can be seen from Fig. \ref{Tuning-Properties-Feedback-STMD}(a), the STMD responds significantly to objects with velocities ranging from $50$ to $400$ pixels/s (output $>0.2$), and reaches its maximal output at $150$ pixels/s.  Note that the interval $[50,400]$ pixels/s and velocity $150$ pixels/s are referred as preferred velocity range and optimal velocity, respectively. Compared to the STMD, the Feedback STMD peaks at a higher velocity of $350$ pixels/s, and has a much wider range of preferred velocities between $150$ and $800$ pixels/s. This is because the time-delay feedback loop can largely suppress responses to objects with low velocities ($<200$ pixels/s), which finally leads to a significant shift of the preferred velocity range toward the high-velocity side. In Fig. \ref{Tuning-Properties-Feedback-STMD}(b), we can see that both the STMD and Feedback STMD give preference to objects whose widths are lower than $20$ pixels, and reach maximum at width $= 10$ pixels. Moreover, these two models have the same preferred height range ($<10$ pixels) as well as the same optimal height ($5$ pixels), as displayed in Fig. \ref{Tuning-Properties-Feedback-STMD}(c). It can be observed from Fig. \ref{Tuning-Properties-Feedback-STMD}(d) that the two models yield higher outputs with the increase of Weber contrast, and finally peak at Weber contrast $=1$. 

In the above experiment, the model with and without feedback all demonstrate velocity selectivity, width selectivity, height selectivity, and Weber contrast sensitivity. From another perspective, the results also indicate that the time-delay feedback cannot directly affect the presence or absence of the four tuning properties, though it actually changes the optimal velocity and preferred velocity range. 

\subsection{Parameter Sensitivity Analysis}
\label{Parameter-Sensitivity-Analysis}

\begin{table*}[t!]
	\renewcommand{\arraystretch}{1.3}
	\caption{Details of the Synthetic Image Sequences that are Categorized into Six Groups in Terms of Six Different Image Parameters.}
	\label{Parameter-Seeting-of-Image-Sequence}
	\centering
	\begin{tabular}{|c|c|c|c|c|c|c|c|}
		\hline
		Image parameter & Initial sequence  & Group 1 & Group 2 & Group 3 & Group 4 & Group 5 & Group 6 \\
		\hline
		Target size ($\text{pixels} \times \text{pixels}$) & $5 \times 5$  & $\mathbf{1 \times 1 \sim 15 \times 15}$ & $5 \times 5$ & $5 \times 5$ & $5 \times 5$ & $5 \times 5$ & $5 \times 5$ \\
		\hline
		Target luminance & $0$ & $0$ & $\mathbf{0\sim 75}$ & $0$  & $0$  & $0$ & $0$ \\
		\hline
		Target velocity (pixels/s)& 250 & 250 & 250  & $\mathbf{0 \sim 500}$& 250 & 250 & $\mathbf{200 \sim 400}$\\	
		\hline
		Background velocity (pixels/s) & $150$ & $150$ & $150$ & $150$ & $\mathbf{0\sim 500}$ & $\mathbf{0\sim 500}$ & $150$ \\
		\hline
		Background motion direction &rightward & rightward  & rightward & rightward & rightward  & \bf{leftward} & rightward \\
		\hline
		Background Image &  Fig.\ref{Input-Image-Frame-Middle-Line-Highlighted} &  Fig.\ref{Input-Image-Frame-Middle-Line-Highlighted} & Fig.\ref{Input-Image-Frame-Middle-Line-Highlighted} & Fig.\ref{Input-Image-Frame-Middle-Line-Highlighted} & Fig.\ref{Input-Image-Frame-Middle-Line-Highlighted} & Fig.\ref{Input-Image-Frame-Middle-Line-Highlighted} & \bf{Fig.\ref{Detection-Performance-Differnet-Backgrounds}(a) $\sim$ (c)} \\
		\hline
	\end{tabular}
\end{table*}

The time-delay feedback loop is completely determined by three preset parameters, including the feedback constant $\alpha$, the order $n_4$  and time constant $\tau_4$ of the Gamma kernel, as indicated in  (\ref{Feedback-Signal-Delay}). We have conducted the parameter sensitivity study in terms of $\alpha$, $n_4$ and $\tau_4$ to evaluate their effects on the tuning properties of the Feedback STMD model.   

Let us first study the effect of modifying the feedback constant $\alpha$ using the following values: $\{0.1, 0.25, 0.5, 1, 2\}$. Fig. \ref{Tuning-Properties-Different-Feedback-Alpha-Order-Tau}(a) shows the tuning properties of the Feedback STMD model with respect to different values of $\alpha$. As can be seen, the optimal velocity increases from $250$ to $400$ pixels/s as $\alpha$ is increased from $0.1$ to $2.0$, and the preferred velocity range is also extended to a much broader span. However, the width selectivity, height selectivity, and  Weber contrast sensitivity, are little affected by the change in $\alpha$. Note that the optimal height ($5$ pixels) and the preferred height range ($<10$ pixels) remain unchanged, though a larger $\alpha$ will lead to a slight increase in the model outputs for heights greater than $10$ pixels. The reason for the above results is that the parameter $\alpha$ controls the strength of the feedback signal, as illustrated in Fig. \ref{Schematic-Model-Output-Feedback-Signal}. The responses to objects with low velocities will be much weaker when subtracting a stronger feedback signal, which finally causes the shift of the optimal velocity to $400$ pixels/s.

Next, let us consider the tuning properties of the Feedback STMD model in relation to different values of $n_4$ and $\tau_4$, which are tuned within the following values ranges: $\{9, 12, 15, 18, 21\}$ and $\{20, 25, 30, 35, 40\}$, respectively. The experimental results are shown in Fig. \ref{Tuning-Properties-Different-Feedback-Alpha-Order-Tau}(b) and (c). As can be seen, the decrease of $n_4$ and $\tau_4$ shifts the optimal velocity to a higher value, and significantly broadens the preferred velocity range towards higher velocities. However, the other three tuning properties are less sensitive to the changes of $n_4$ and $\tau_4$. This is because $n_4$ and $\tau_4$ control the response duration and time-delay length of the feedback signal, respectively [see Fig. \ref{Schematic-Model-Output-Feedback-Signal}]. Lower values of  $n_4$ (or $\tau_4$) generally result in a stronger feedback signal, and therefore weaker responses to slow-moving objects. 


\begin{figure*}[!t]
	\centering
	\subfloat[]{\includegraphics[width=0.275\textwidth]{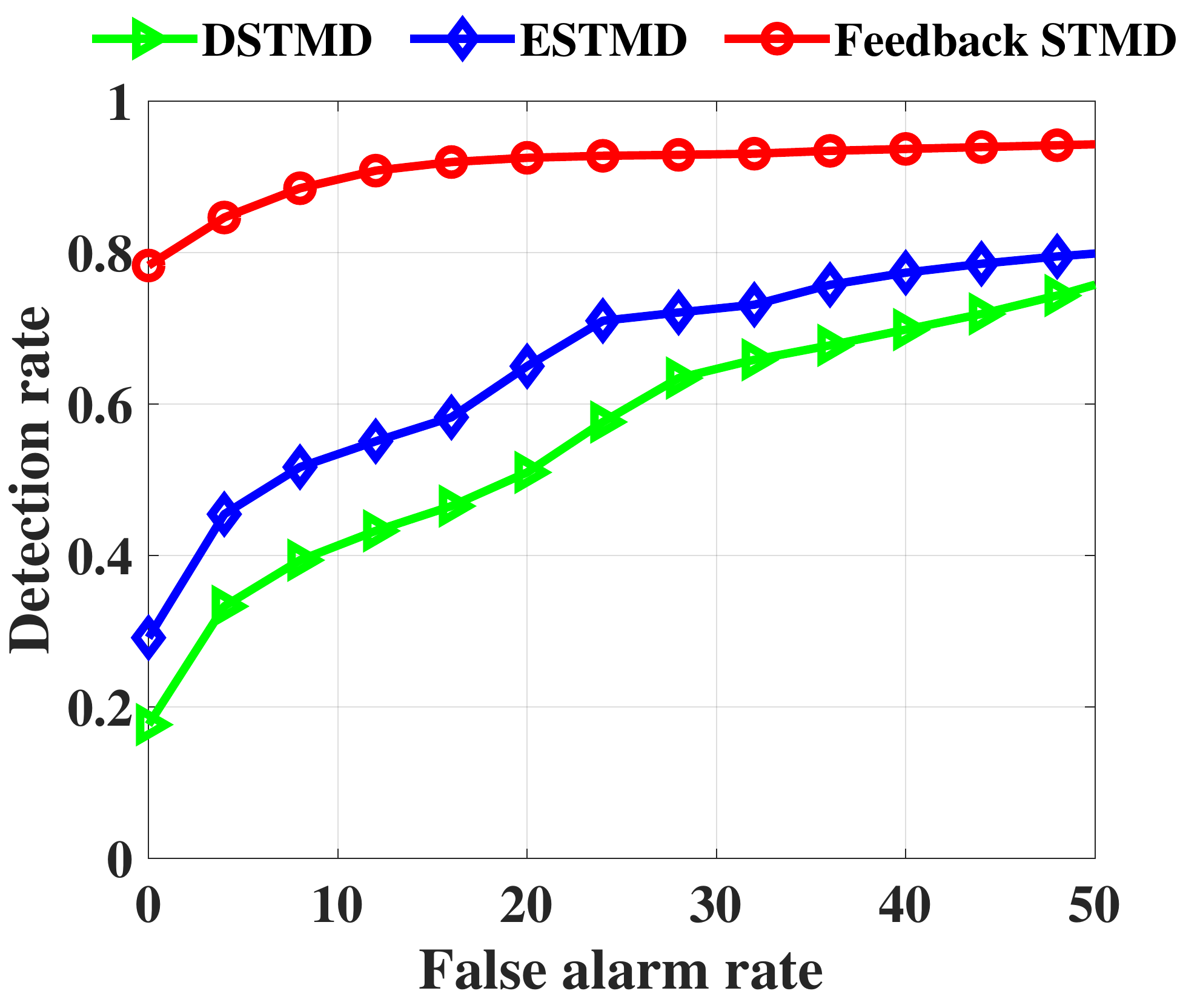}}
	\hfil
	\subfloat[]{\includegraphics[width=0.275\textwidth]{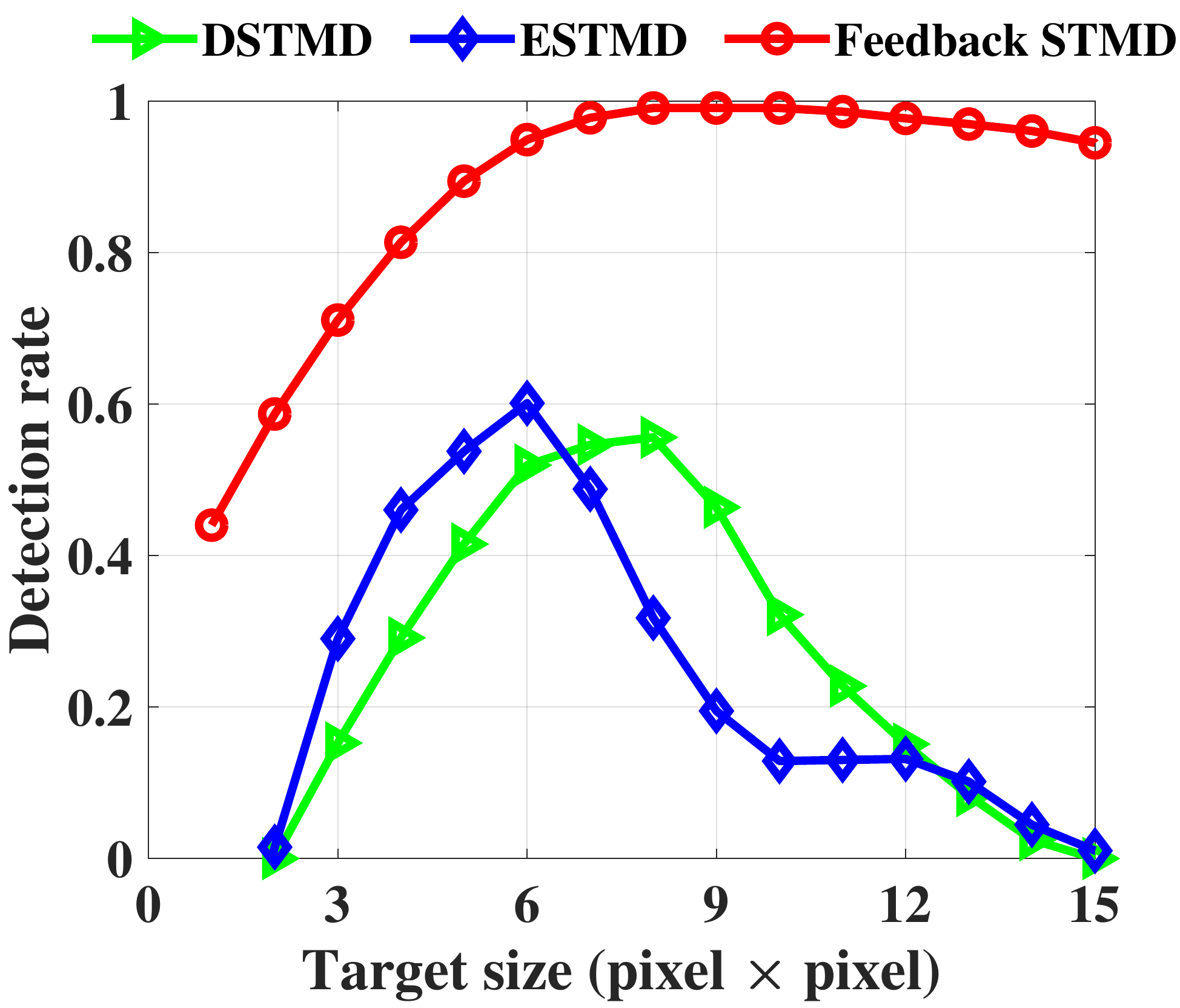}}
	\hfil  
	\subfloat[]{\includegraphics[width=0.275\textwidth]{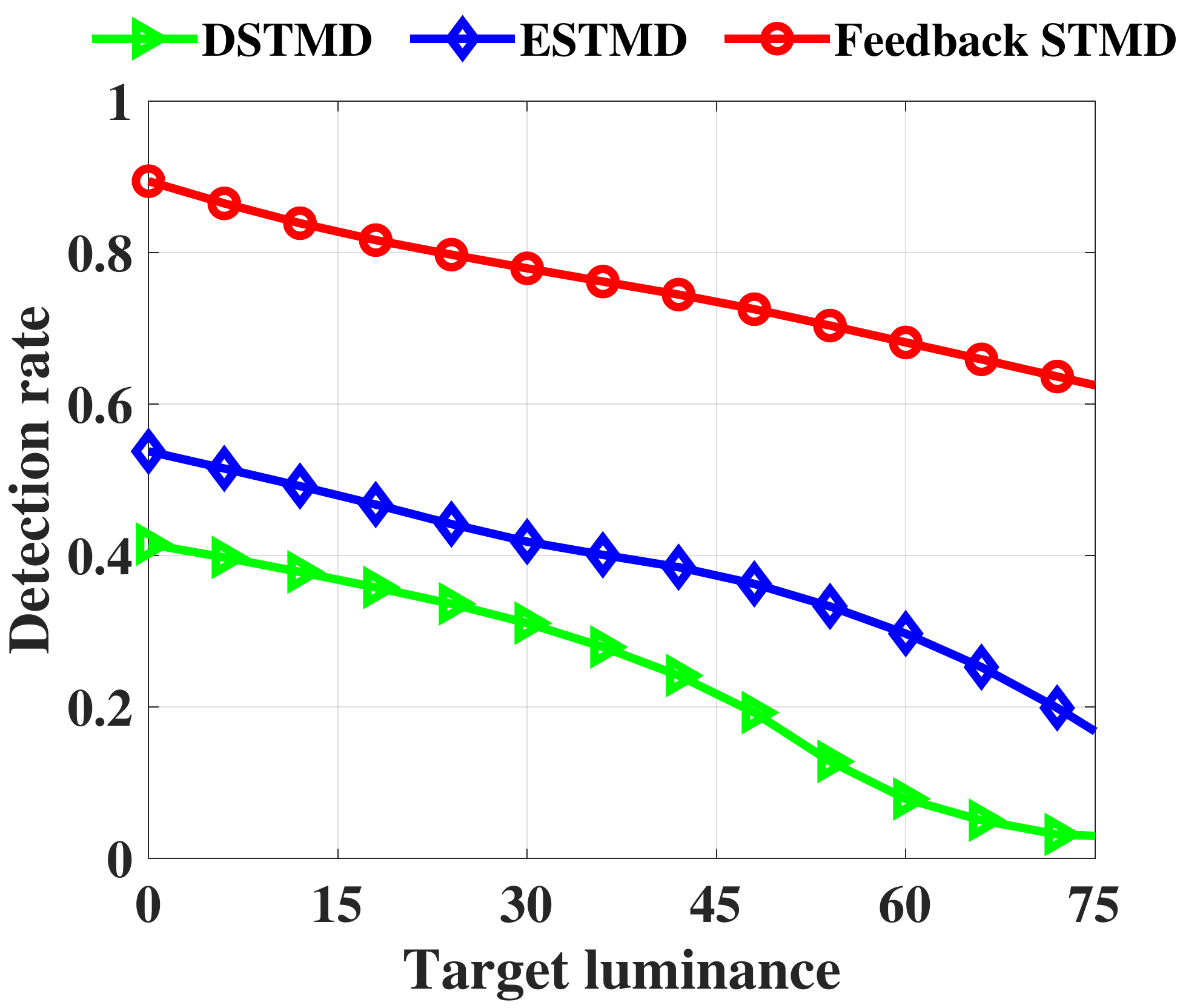}}
	\hfil
	\subfloat[]{\includegraphics[width=0.275\textwidth]{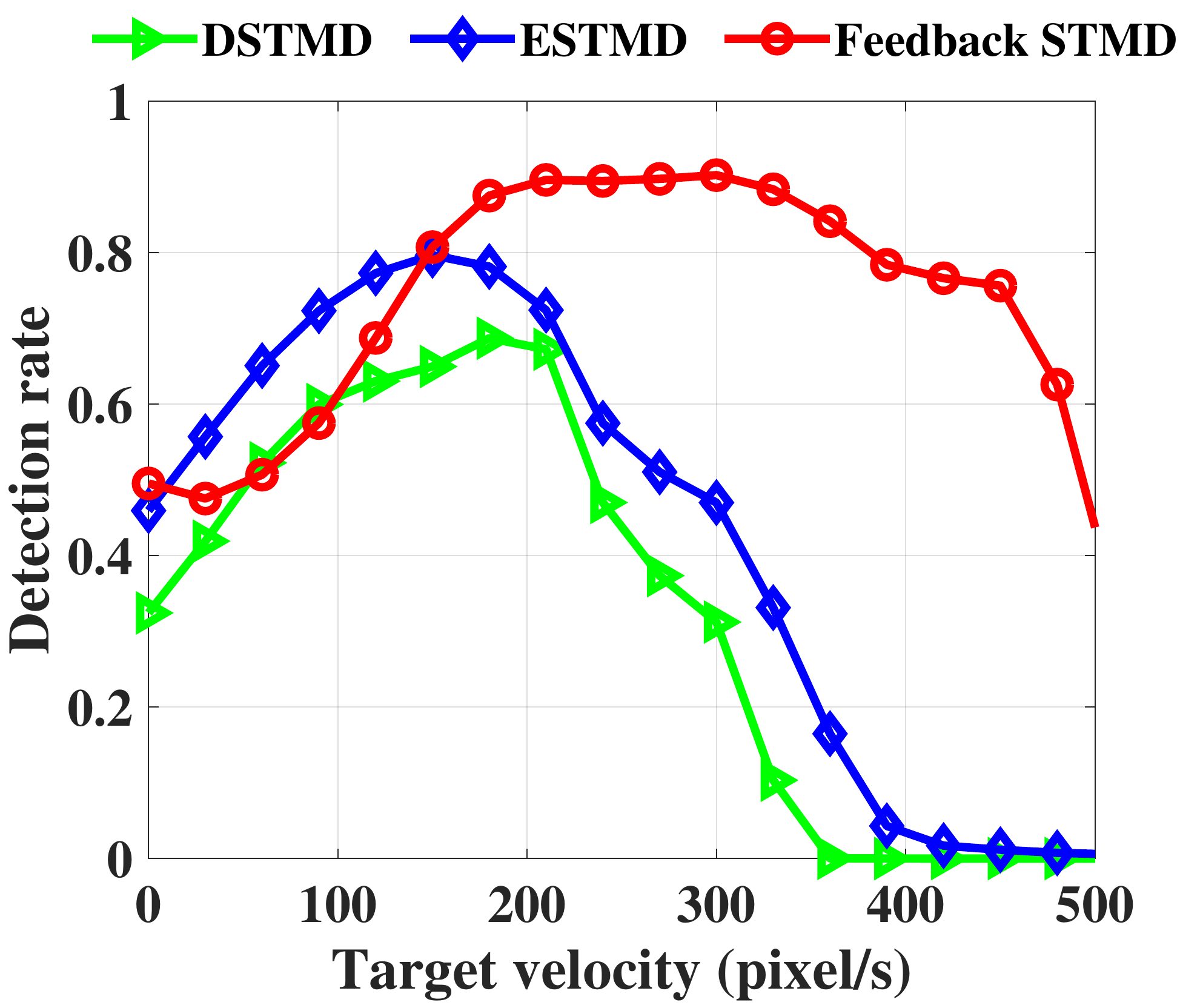}}
	\hfil
	\subfloat[]{\includegraphics[width=0.275\textwidth]{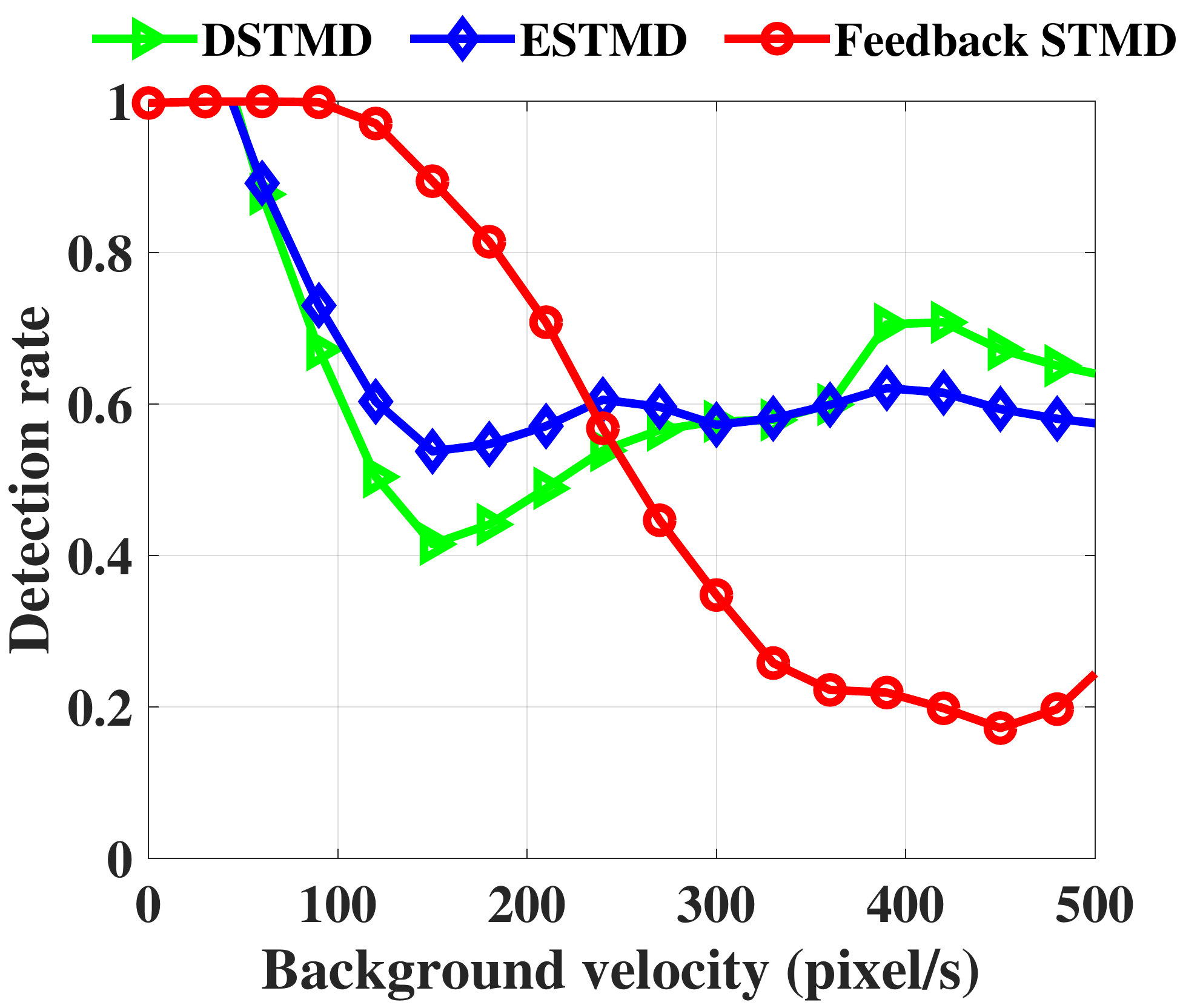}}
	\hfil
	\subfloat[]{\includegraphics[width=0.275\textwidth]{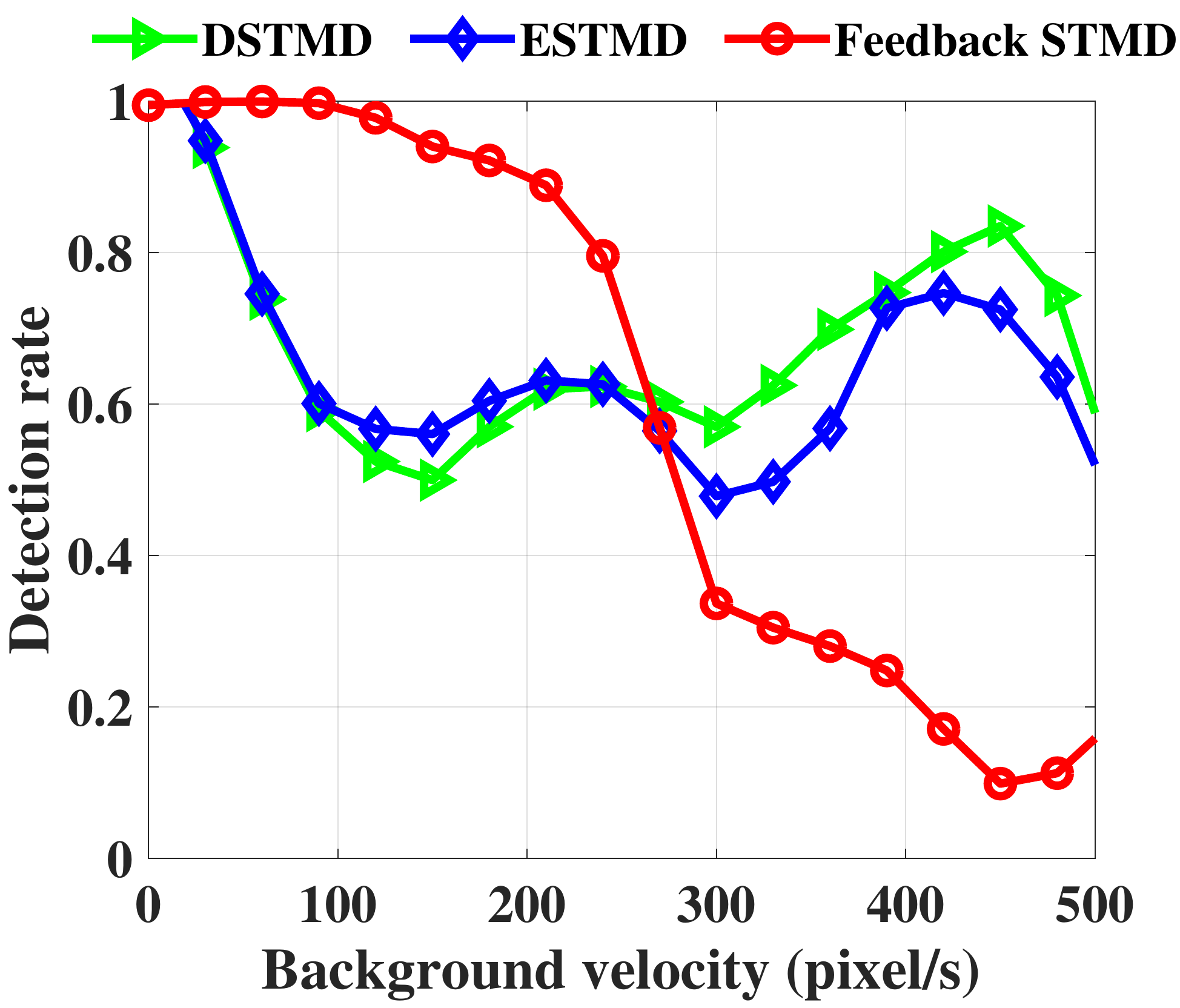}}
	\caption{(a) ROC curves of the Feedback STMD and the two baseline models on the initial image sequence. (b)-(f) Detection rates under the fixed false alarm rate $F_A = 10$ with respect to (b) target size, (c) target luminance, (d) target velocity, (e) background velocity (leftward motion), and (f) background velocity (rightward motion). The Feedback STMD improves detection performance for fast-moving small targets.}
	\label{CB-1-LDTB-Size-Velocity-DR-FA}
\end{figure*}

Hence, the results shown in Fig. \ref{Tuning-Properties-Different-Feedback-Alpha-Order-Tau} reveal a practical approach for optimizing model performance for small target motion detection. By estimating the velocity of the small target is estimated in advance, the parameters of the feedback loop, i.e., $\alpha$, $n_4$, and $\tau_4$, can be tuned so that the appropriate optimal velocity and preferred velocity range of the model can be selected.

\subsection{Comparison on Synthetic and Real Data Sets}

\begin{table}[t]
	\renewcommand{\arraystretch}{1.3}
	\caption{Running time comparison of the three models on the videos in Vision Egg dataset. The value is averaged over all the tested videos, each of which is with $1000$ frames while with frame size of $500$ pixels (horizontal) $\times$ $250$ pixels (vertical).}
	\label{Comparison-of-Running-Time-STMD}
	\centering
	\begin{tabular}{c|c|c|c}
		\hline
		 Method & ESTMD  & DSTMD & Feedback STMD \\
		\hline
		Time (s/frame) & $0.064$  & $0.116$ & $0.084$ \\
		\hline
	\end{tabular}
\end{table}

We have compared the proposed Feedback STMD model with two baseline methods, including  ESTMD \cite{wiederman2008model} and DSTMD \cite{wang2018directionally}, using both synthetic and real data sets in terms of detection rate and false alarm rate. As shown in Table \ref{Parameter-Seeting-of-Image-Sequence}, the synthetic image sequences are classified into six groups based on target size, target luminance, target velocity, background velocity, background motion direction, and background images. This has allowed us to study the relationships between model performance and these six image parameters. To achieve meaningful comparisons, the baseline models have been appropriately tuned to cover the same preferred velocity range and size range with the STMD [see Fig. \ref{Tuning-Properties-Feedback-STMD}].

\begin{figure*}[!t]
	\centering
	\subfloat[]{\includegraphics[width=0.285\textwidth]{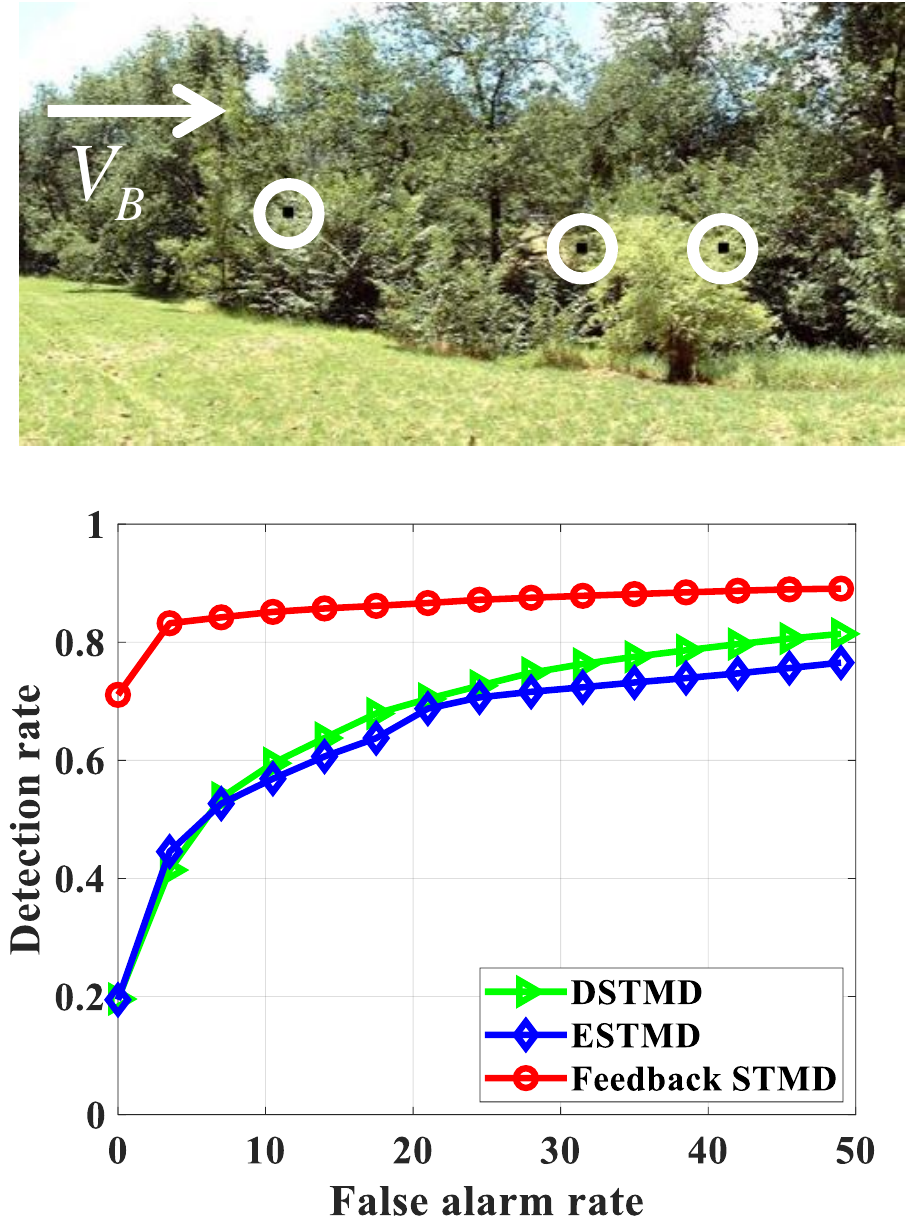}}
	\hfil
	\subfloat[]{\includegraphics[width=0.285\textwidth]{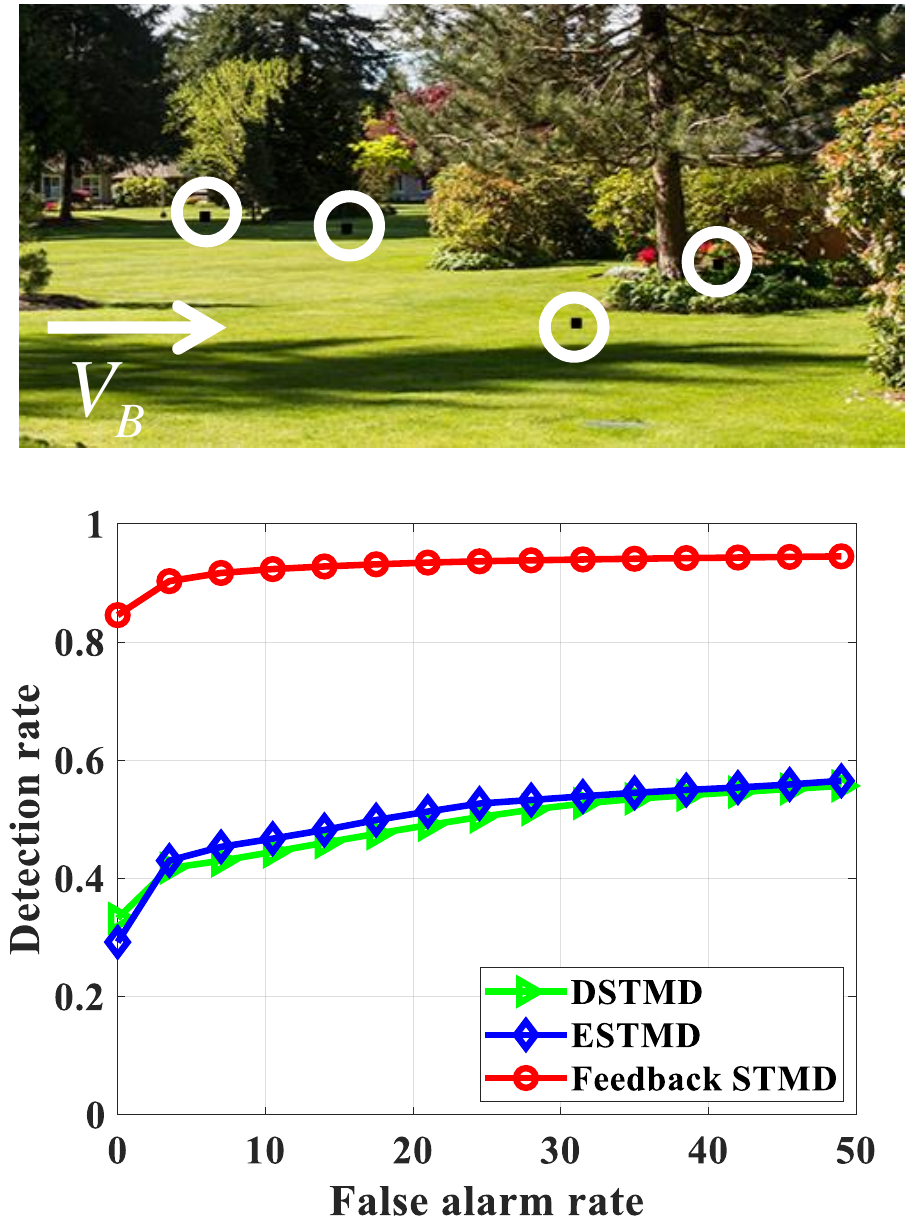}}
	\hfil
	\subfloat[]{\includegraphics[width=0.285\textwidth]{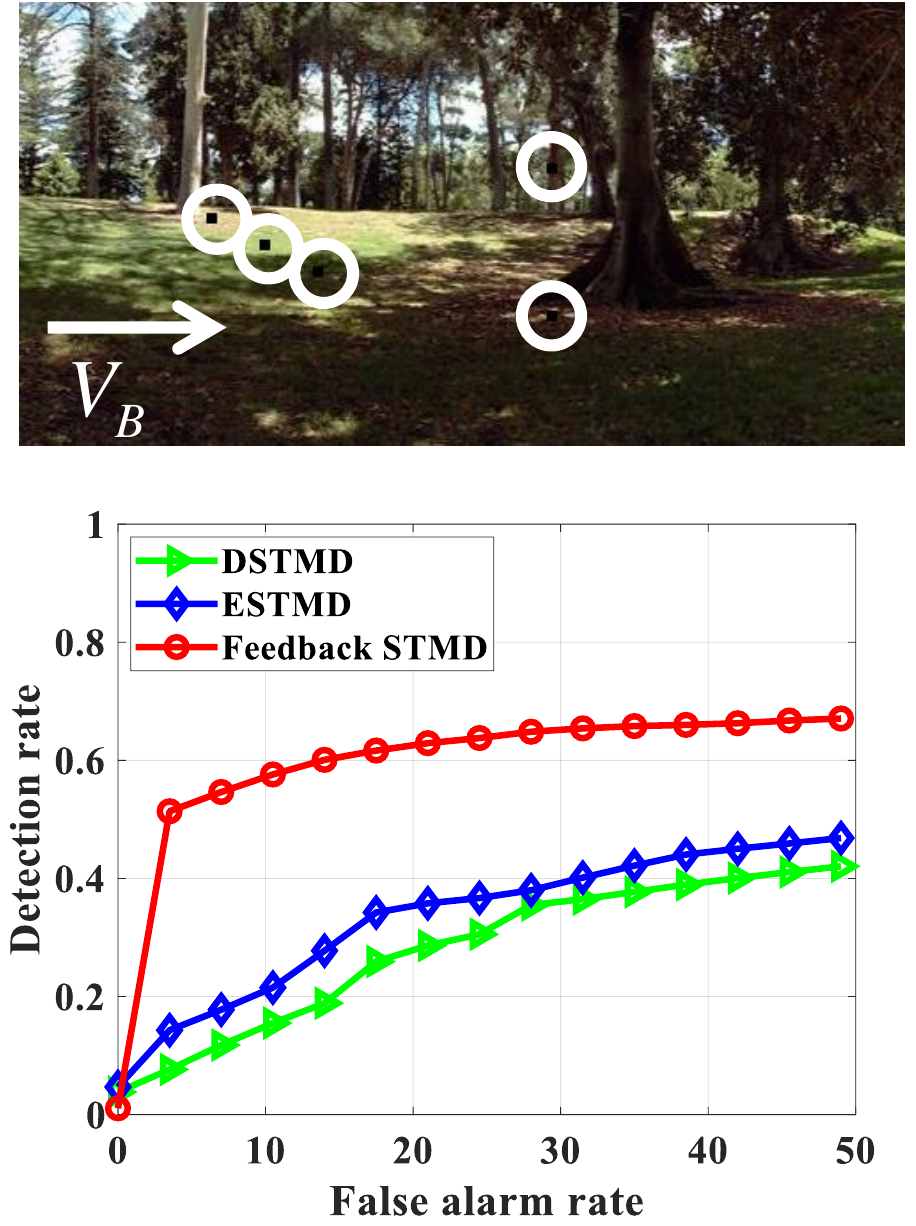}}
	\caption{Representative frames of the three videos that hold multiple small moving targets against cluttered background, and ROC curves of the proposed Feedback STMD model and the two baseline models on the three videos are shown in (a)-(c), respectively. The Feedback STMD outperforms the other two models for  multiple small target detection against different background images.}
	\label{Detection-Performance-Differnet-Backgrounds}
\end{figure*}


\begin{figure*}[t!]
	\centering
	\subfloat[]{\includegraphics[width=0.275\textwidth]{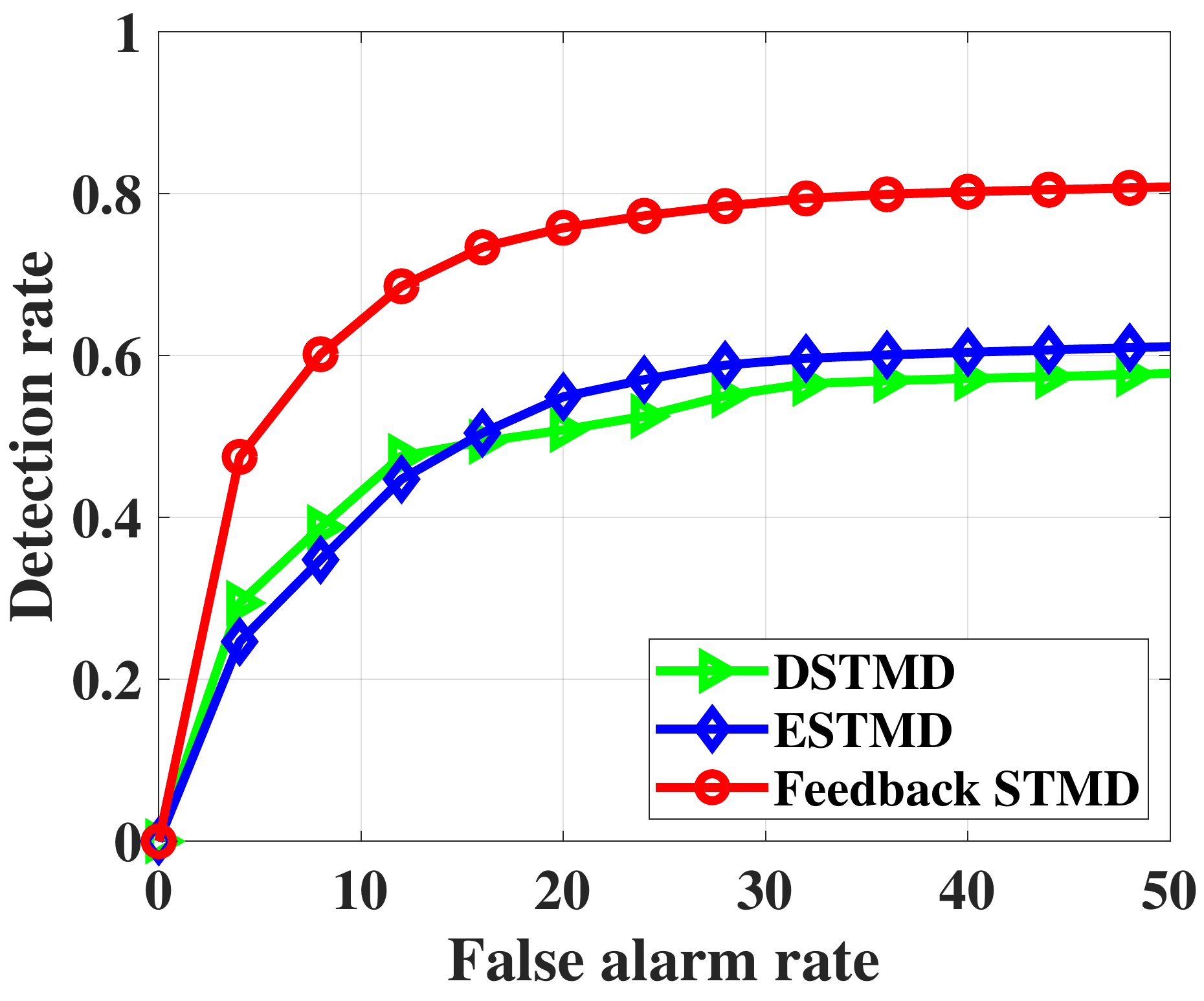}}
	\hfil
	\subfloat[]{\includegraphics[width=0.275\textwidth]{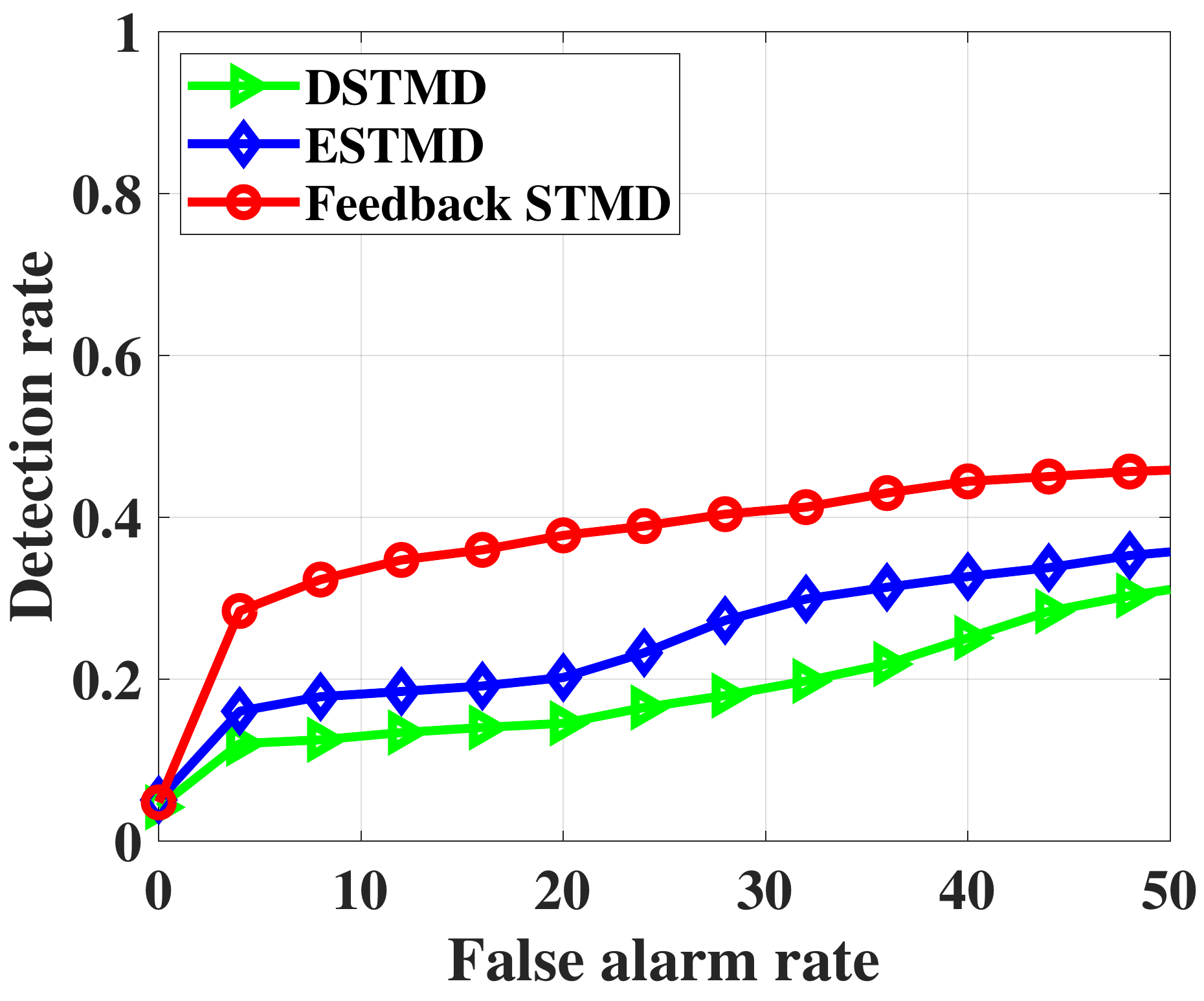}}
	\hfil
	\subfloat[]{\includegraphics[width=0.275\textwidth]{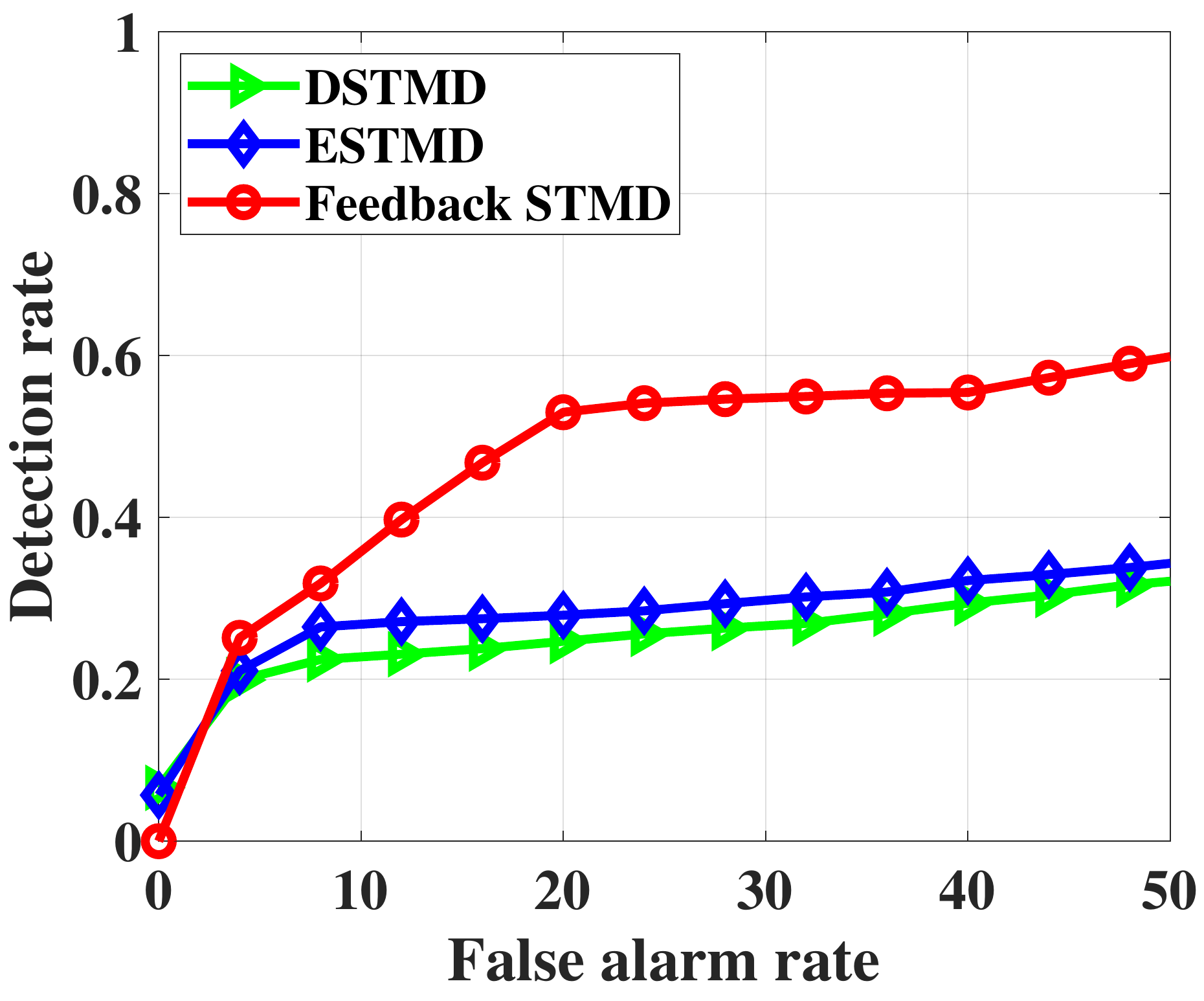}}
	\hfil
	\subfloat[]{\includegraphics[width=0.275\textwidth]{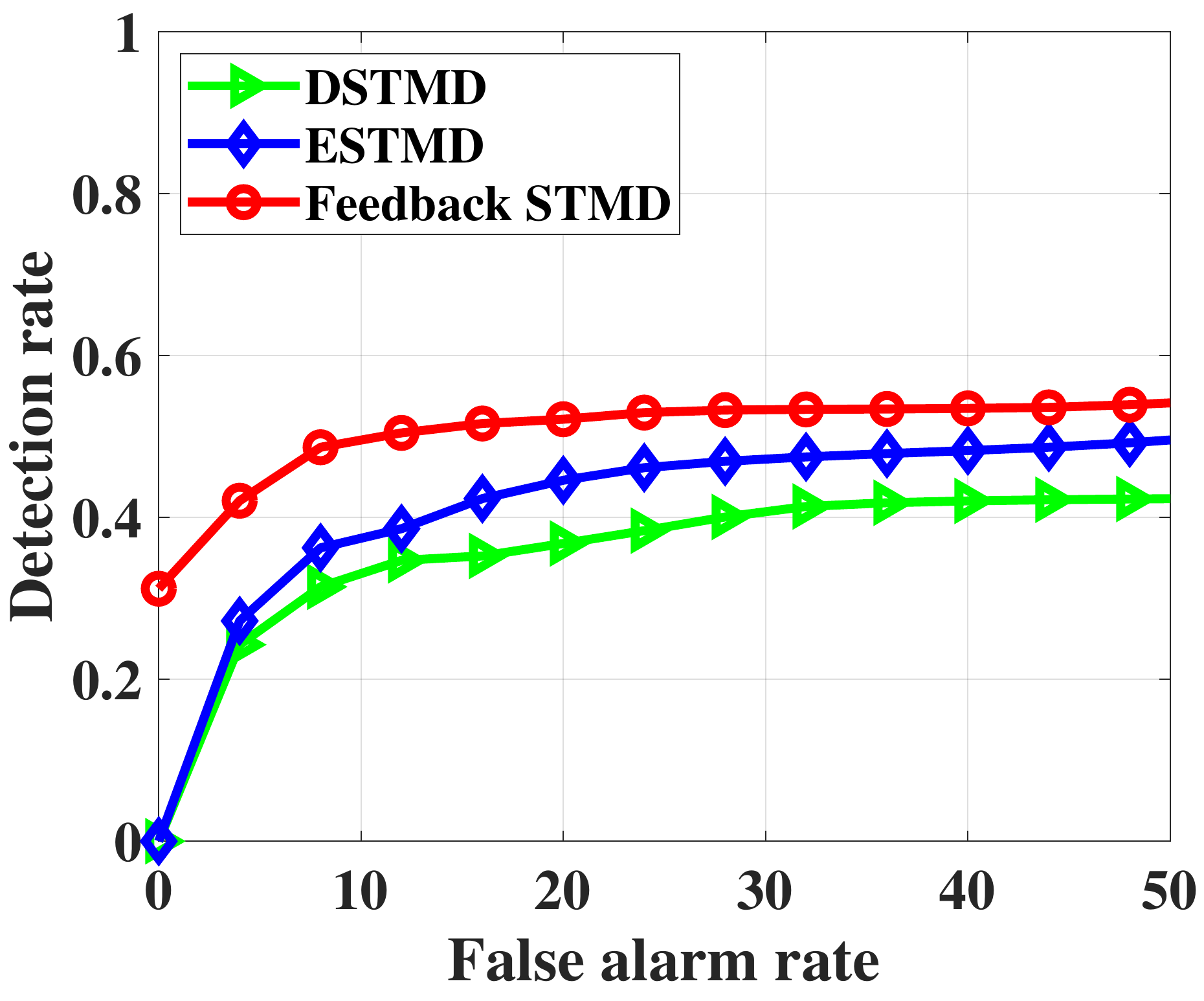}}
	\hfil
	\subfloat[]{\includegraphics[width=0.275\textwidth]{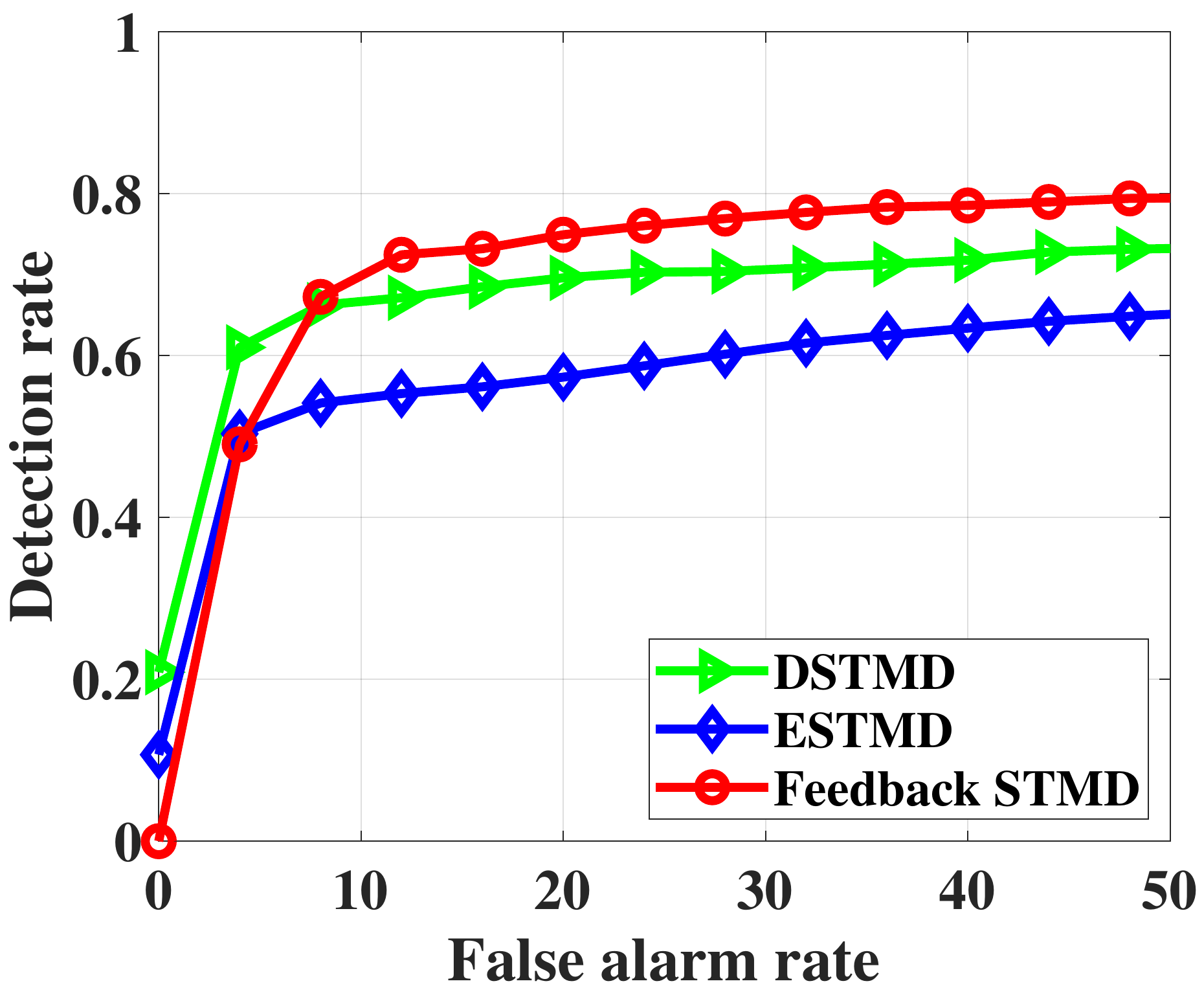}}
	\hfil
	\subfloat[]{\includegraphics[width=0.275\textwidth]{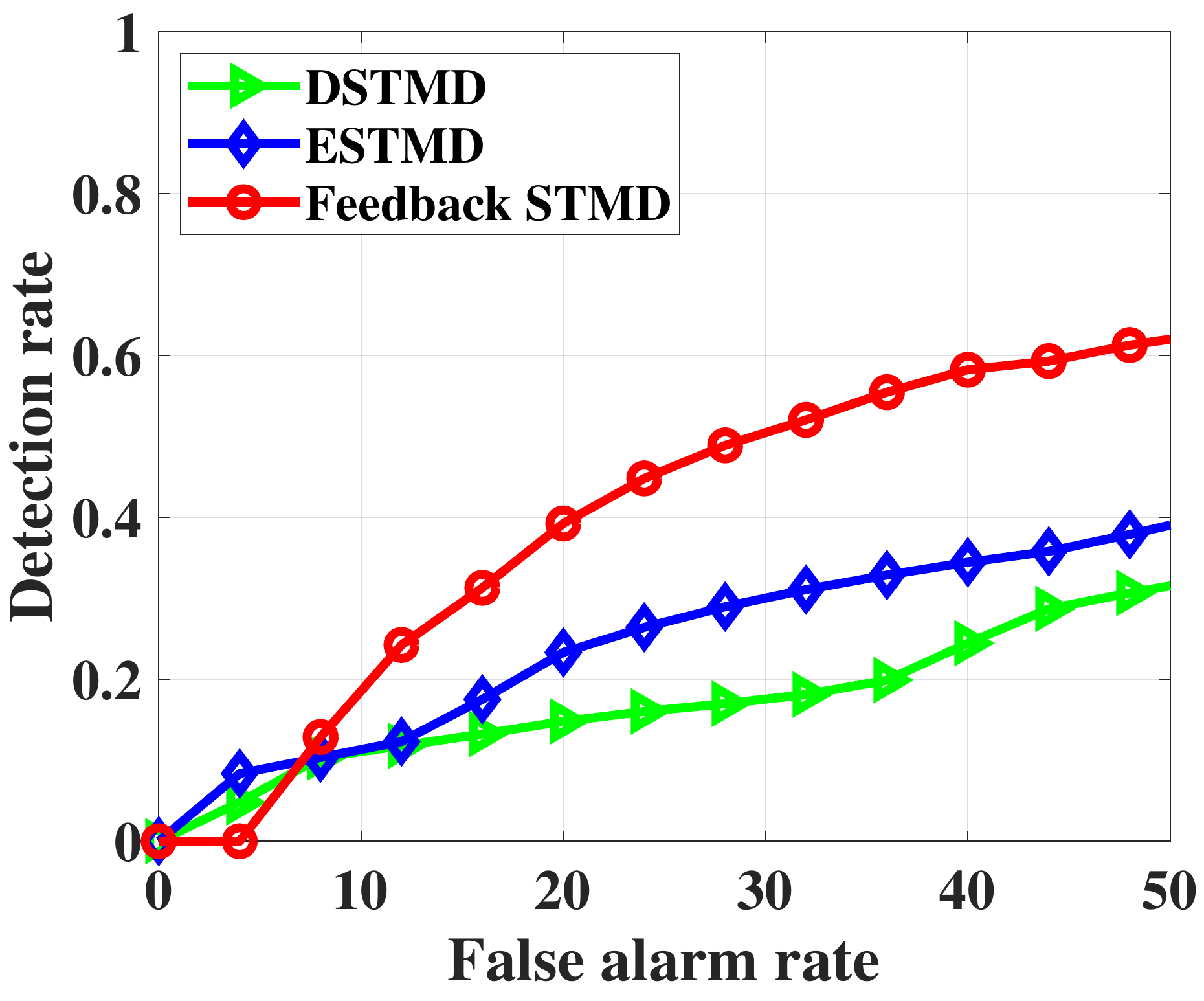}}
	
	\caption{ROC curves of the proposed Feedback STMD on six real videos in comparison to the ESTMD and DSTMD models. (a) Real video $1$ (GX010071). (b) Real video $2$ (GX010250). (c) Real video $3$ (GX010303). (d) Real video $4$ (GX010307). (e) Real video $5$ (GX010322). (f) Real video $6$ (GX010337). The Feedback STMD model achieves better performance than the other two models on all real videos.}
	\label{Real-Dataset-ROC-Curve}
\end{figure*}

The receiver operating characteristics (ROC) curves created from the initial image sequence by the proposed Feedback STMD model and the two baseline models are shown in Fig. \ref{CB-1-LDTB-Size-Velocity-DR-FA}(a). It is clear that the detection performance of the Feedback STMD model  far-exceeds that of either baseline model. The detection rate of the Feedback STMD model is consistently higher than the rates of the ESTMD and DSTMD models at any false alarm rate. In addition, Fig. \ref{CB-1-LDTB-Size-Velocity-DR-FA}(b)-(f) display the detection rates of the three models for the Group $1$-$5$, respectively, where the false alarm rate is set at $10$. When varying target size, the Feedback STMD model achieves the best performance of the three models, as shown in Fig. \ref{CB-1-LDTB-Size-Velocity-DR-FA}(b). The detection rate of the Feedback STMD model remains at a high value ($>0.9$) when the target size increases from $6 \times 6$ to $15 \times 15$ pixels. However, the two baseline models both experience a sharp decrease in detection rates to $0$, after reaching their maxima at a target size about $ 6\times 6$ pixels. Likewise for different levels of target luminance, the Feedback STMD model consistently outperforms the ESTMD and DSTMD models, as displayed in Fig. \ref{CB-1-LDTB-Size-Velocity-DR-FA}(c). The results also show that an increase in target luminance results in performance degradation of all three models. Regarding the impact of target velocity, the results in Fig. \ref{CB-1-LDTB-Size-Velocity-DR-FA}(d) demonstrate that as velocity is increased to about $150$ pixels/s, all three models show similar detection rates, which increase from about $0.4$ to $0.8$. However, as target velocity is increased beyond this point, the detection rates of the two baseline models rapidly fall off whilst, in contrast, the Feedback STMD model maintains a much superior, high detection rate up to target velocities in excess of $400$ pixels/s. At this point, the detection rates for both baseline models have fallen to zero. Hence, for the Feedback STMD model, a greater target velocity enables small targets to be readily discriminated from the background. The results for the impact of background velocity are presented in Fig. \ref{CB-1-LDTB-Size-Velocity-DR-FA}(e) and (f). They both show that in comparison to the baseline models, the Feedback STMD model significantly improves detection rates when the background moves more slowly than the small target ($<250$ pixels/s). However, it performs much worse than the baseline models when the background velocity is higher than that of the small target. The reason for this is that the time-delay feedback makes the Feedback STMD model better at  detecting fast-moving small targets but not objects moving with low velocities. When the background is moving faster than the small target ($> 250$ pixels/s), the background false positives will receive much weaker suppression from the feedback loop, which consequently leads to the decrease in detection rate.

The capabilities of the models to discern multiple small moving targets against different backgrounds are reported in Fig. \ref{Detection-Performance-Differnet-Backgrounds}. The results show that the ROC curves of the proposed Feedback STMD model and the baseline models for three videos (the Group $6$). It can be observed that the Feedback STMD model performs much better than either the ESTMD and DSTMD models in detecting multiple small targets against all background images. 

We have further evaluated the models on six real videos and report their ROC curves in Fig. \ref{Real-Dataset-ROC-Curve}. The videos are randomly selected from the RIST data set, each of which displays a fast-moving small target against the cluttered background. The corresponding video numbers are given in the caption. As can be seen, the Feedback STMD model outperforms the ESTMD and DSTMD models on all six videos. Specifically, its detection rate is always higher than those of the other two models at any false alarm rate.

We have also compared the running time of the three models. As can be seen from Table \ref{Comparison-of-Running-Time-STMD}, the proposed Feedback STMD model needs $0.084$s to process a frame, which is comparable to the ESTMD model, and is more efficient than the DSTMD model.

\section{Conclusion}
\label{Conclusion}

In this paper, we have developed an STMD-based neural network with time-delay feedback (Feedback STMD) to discriminate fast-moving small targets from cluttered backgrounds. The proposed model contains four sequentially arranged neural layers and a time-delay feedback loop. The four neural layers, interconnected by feedforward connections are intended to extract motion information about small targets by computing luminance change of each pixel with respect to time. The feedback loop is designed to propagate the extracted motion information to lower neural layers to inhibit slow-moving background false positives. The model with and without feedback were evaluated and compared on  synthetic and real data sets to demonstrate the effectiveness of feedback. Experimental results show that the time-delay feedback can maintain model responses to fast-moving objects, while significantly suppressing those with lower velocities. Moreover, it is able to improve detection performance for small targets with velocities higher than that of the complex background. In the future, we will consider other possible feedback types such as time-varying feedback, and explore their self-adaptability for various objects and environments.

\ifCLASSOPTIONcaptionsoff
  \newpage
\fi


\bibliographystyle{IEEEtran}

\bibliography{IEEEabrv,Reference}

\end{document}